\pdfoutput=1
\documentclass[11pt]{article}

\usepackage[preprint]{acl}

\usepackage{times}
\usepackage{latexsym}

\usepackage[T1]{fontenc}

\usepackage[utf8]{inputenc}

\usepackage{microtype}

\usepackage{inconsolata}

\usepackage{graphicx}

\usepackage{multirow}
\usepackage{subcaption}
\usepackage{listings}
\usepackage{color}

\usepackage{amsmath} 
\usepackage{fvextra}
\usepackage{makecell} 

\definecolor{codegreen}{rgb}{0,0.6,0} 

\title{DSAI: Unbiased and Interpretable Latent Feature Extraction for Data-Centric AI}

\author{
 \textbf{Hyowon Cho\textsuperscript{1, \thanks{Work done while interning at NAVER AI Platform}}},
 \textbf{Soonwon Ka \textsuperscript{2}},
 \textbf{Daechul Park\textsuperscript{2}},
 \textbf{Jaewook Kang\textsuperscript{2}}, 
 \textbf{Minjoon Seo\textsuperscript{1,\thanks{Corresponding authors.}}},
 \newcommand\CoAuthorMark{\footnotemark[\arabic{footnote}]}
 \textbf{Bokyung Son\textsuperscript{2,\protect\CoAuthorMark}}
\\
 \textsuperscript{1}KAIST AI, 
 \textsuperscript{2} NAVER AI Platform,
\\
\small{
   \texttt{\textsuperscript{1}\{hyyoka, minjoon\} @kaist.ac.kr}
 }\\
 \small{
   \texttt{\textsuperscript{2}\{soonwon.ka, daechul.park, jaewook.kang, bo.son\} @navercorp.com}
 }
}

\begin{document}
\maketitle
\begin{abstract}
Large language models (LLMs) often struggle to objectively identify latent characteristics in large datasets due to their reliance on pre-trained knowledge rather than actual data patterns. To address this data grounding issue, we propose Data Scientist AI (DSAI), a framework that enables unbiased and interpretable feature extraction through a multi-stage pipeline with quantifiable prominence metrics for evaluating extracted features. On synthetic datasets with known ground-truth features, DSAI demonstrates high recall in identifying expert-defined features while faithfully reflecting the underlying data. Applications on real-world datasets illustrate the framework's practical utility in uncovering meaningful patterns with minimal expert oversight, supporting use cases such as interpretable classification \footnote{The title of our paper is chosen from multiple candidates based on DSAI-generated criteria.}.
\end{abstract}

\section{Introduction}
\vspace{-1mm}

The ability to analyze large-scale datasets is a cornerstone of deriving actionable business insights. Traditionally, this task has been managed by human \textit{data scientists}, but it faces several key challenges: (1) the large volume of data makes it difficult to review all information comprehensively, (2) human analysis can often include subjective bias, and (3) collaboration with domain experts is often required, leading to high operational costs.

Large language models (LLMs) have emerged as powerful tools for identifying patterns within massive datasets, leveraging their ability to process and generate language in context \cite{touvron2023llama, dubey2024llama, achiam2023gpt, gpt4o, lloom, tnt-llm}. However, their application to data analysis is limited by critical shortcomings. First, LLMs often struggle to identify latent characteristic patterns in 
\vspace{-1mm}
large datasets due to inherent \textit{data grounding issues}, where outputs rely on pre-trained knowledge rather than the specific nuances of the input data \cite{kossen2024context, kenthapadi2024grounding, wu2024clasheval}. Second, the difficulty in verifying LLM-generated responses and the lack of quantitative evaluation methods require expert oversight, which can be prohibitively expensive at scale.



To address these limitations, we propose \textbf{Data Scientist AI (DSAI)}, a framework that systematically applies LLMs to extract and refine latent features from data. Unlike direct feature extraction approaches, DSAI adopts a bottom-up approach, starting with detailed analysis of individual data points, aggregating their characteristics, and deriving actionable features. 
The process is guided by defined \textit{perspectives} which provide LLMs with a consistent framework for interpreting data points while minimizing subjective bias.


The DSAI pipeline operates in five stages:
\textbf{\#1 Perspective Generation} identifies data-driven perspectives from a small subset of data.
\textbf{\#2 Perspective-Value Matching} assigns values to individual data points by evaluating them against these perspectives.
\textbf{\#3 Clustering} groups values with shared characteristics to reduce redundancy.
\textbf{\#4 Verbalization} converts extracted features into a compact criterion form.
\textbf{\#5 Prominence-based Selection} determines which features to use based on a prominence intensity metric that quantifies the discriminative power of each extracted feature.

Throughout these stages, the LLM remains task-agnostic – we do not reveal the specific domain or the “correct answer” context during feature generation. This design minimizes bias and ensures that the identified features are grounded in the data rather than the model’s background knowledge.

We validate our framework on datasets curated for our experiments, including a research titles dataset \cite{dataset_title} and an advertising slogans dataset \cite{dataset_slogan}, each with expert-defined criteria that serve as ground truth for evaluation. We then demonstrate DSAI’s practical value on three real-world datasets: news headlines with click-through rate (CTR) labels \cite{dataset_mind}, a spam detection dataset \cite{dataset_spam}, and Reddit comments with community engagement metrics \cite{dataset_reddit}.

Our main contributions are as follows: 
\begin{itemize}
    \item \textbf{Minimizing Bias}: We ensure that LLMs focus on latent characteristics present in the data, thus mitigating the tendency to rely on domain-specific prior knowledge (addressing the data grounding issue).
    \item \textbf{Prominence Metric}: We introduce a quantitative metric for feature prominence, which serves as a proxy for the discriminative power of each extracted feature.
    \item \textbf{Interpretability}: We improve interpretability through feature-to-source traceability, allowing users to trace each extracted feature back to the data points that support it.
    \item \textbf{Efficiency}: We enable thorough examination of large datasets with minimal human labor by systematically guiding LLMs through the analysis process.

\end{itemize}

\vspace{-1mm}
\section{Related Works}
\vspace{-1mm}
\subsection{Latent Feature Extraction with LLM}
\vspace{-1mm}

Recent advancements in LLMs have demonstrated their effectiveness in extracting latent features, particularly in identifying perspectives and matching values in data \cite{cotam}. Studies using LLM-based clustering techniques  have shown promising results in extracting high-level concepts\cite{lloom, tnt-llm, viswanathan-etal-2024-large, pham2024topicgptpromptbasedtopicmodeling}, demonstrating their utility for analyzing large datasets \cite{goalex, ictc}. 
Research has also shown that decomposing complex tasks into multiple stages or aspects enhances performance \cite{BSM, HD-Eval}. 
Our framework combines perspective generation, multi-stage feature construction, and clustering using LLMs to conduct comprehensive latent feature extraction.

\vspace{-1mm}
\subsection{Bias of LLMs}
\vspace{-1mm}


LLMs face challenges in adapting to new patterns due to pre-existing knowledge biases. \cite{kossen2024context} show that in-context learning (ICL) struggles to overcome these biases even with explicit prompts or many-shot examples. Using external knowledge bases also has limited effectiveness in reducing hallucinations and reliance on internal biases  \cite{kenthapadi2024grounding, lee2023well}. Advanced LLMs fail to align with provided context in about 40\% of predictions when the given context conflicts with their prior knowledge \cite{wu2024clasheval}. 
These findings emphasize the need for robust techniques to mitigate bias and enhance grounding, especially for applications where data-driven conclusions are crucial.

\vspace{-1mm}
\section{Challenges in LLM-Driven Data Analysis}

In this section, we examine the behavior of a state-of-the-art LLM when tasked with directly generating features from data. These exploratory experiments reveal the data-grounding challenges that motivate our DSAI approach.

\subsection{Setting}
\label{llm-driven-data-analysis}
\paragraph{Dataset Annotation and Sampling}

We use two expert-annotated text datasets where the “ground truth” latent features are known (defined by domain experts). The first dataset consists of research paper titles and the second contains advertising slogans. In each dataset, experts \cite{title_expert1, title_expert2, slogan_expert1, slogan_expert2} have outlined a set of specific criteria or features that good examples should exhibit. We annotated each sample for the presence or absence of each criterion. This yields a detailed profile of which features are present in each data point. We then designated \textit{positive} examples as those that satisfy many of the expert criteria (top-scoring "high-quality” samples) and \textit{negative} examples as those that violate or lack many of the criteria (bottom-scoring "low-quality” samples). This dataset construction emphasizes concrete feature differences rather than a subjective quality judgment.

\paragraph{Model}

All experiments in this section use GPT-4o \cite{gpt4o} as the LLM, which is shown to have strong capabilities in understanding nuanced text and performing annotation-like tasks \cite{tan2024largelanguagemodelsdata}. The model was prompted in a zero-shot manner to generate dataset features under various conditions.

\subsection{Data Grounding Issues}
\label{llm_data_grounding_issue}

\begin{table}[t]
\fontsize{10}{12}\selectfont
\centering
\resizebox{\columnwidth}{!}{
\begin{tabular}{c|c|r|r}
\Xhline{1.5pt}
\multirow{2}{*}{\textbf{Reference Criteria}} & \multirow{2}{*}{\textbf{Comparison Criteria}} & \multicolumn{2}{c}{\textbf{Recall (\%)}} \\\cline{3-4}
& & \textbf{Slogan} & \textbf{Title} \\
\Xhline{1.5pt}
\textsc{FlippedPosData} & \textsc{PosData} & 89.5\% & 94.1\% \\
\hline
\textsc{FlippedMixedData} & \textsc{MixedData} & 81.8\%  & 89.5\% \\
\hline
\multirow{4}{*}{\textsc{NoData}} & \textsc{PosData} & 100.0\% &  95.6\%\\
& \textsc{MixedData} & 100.0\% & 92.3\%\\
& \textsc{FlippedPosData} &  90.0\% & 94.1\%\\
& \textsc{FlippedMixedData} & 96.4\% & 85.2\%\\
\hline
\multirow{8}{*}{\textsc{Expert}} & \textsc{NoData} & 88.9\% & 83.3\% \\
& \textsc{PosData} & 66.7\% & 54.2\% \\
& \textsc{MixedData} & 50.0\% & 75.0\% \\
& \textsc{FlippedPosData} & 94.4\% & 45.8\% \\
& \textsc{FlippedMixedData} & 77.8\% & 83.3\% \\ 
\cline{2-4}
& \textsc{\textbf{DSAI (Thres: [0])}} & 100.0\% & 83.3\% \\
& \textsc{\textbf{DSAI (Thres: [0.348])}} & 88.9\% & 83.3\% \\
& \textsc{\textbf{DSAI (Thres: [0.692])}} & 77.8\% & 75.0\% \\
\Xhline{1.5pt}
\end{tabular}}
\caption{Recall of features derived from different approaches across slogans and titles.}
\vspace{-3mm}
\label{expert_vs_baselines}
\end{table}

Using the above datasets, we explored straightforward ways of prompting the LLM to extract latent features, and evaluated whether the LLM’s outputs were truly grounded in the input data. We tried two input configurations for the prompt:
\begin{itemize}
\vspace{-1mm}
    \item \textbf{\textsc{PosData}}: Provide the LLM with only positive examples and ask it to identify characteristics common to these examples. 
\vspace{-1mm}
    \item \textbf{\textsc{MixedData}}: Provide the LLM with both positive and negative examples, and ask it to find distinguishing features. 
\end{itemize}

\vspace{-1mm}

In both cases, we formatted the prompt to list a set of representative samples and requested the model to output a list of key features or criteria describing the positive group. We found that both \textsc{PosData} and \textsc{MixedData} prompts yielded feature lists that substantially overlapped with the expert-defined criteria (Table \ref{expert_vs_baselines}). On the surface, this suggests the model can produce reasonable-sounding features. However, because our datasets come from well-known domains, we must question whether the LLM actually derived these features from the given data, or if it simply regurgitated its own prior knowledge about the domain. To investigate this, we designed four experiments around the following questions:

\paragraph{(a) Does the LLM Adapt to Input Data?}
We tested whether the LLM truly uses the input examples to adapt its feature generation. If the model is grounding its output in the provided data, then changing the data labels should lead to corresponding changes in the generated features. To check this, we flipped the class labels in the input and observed the effect on the output features:

\begin{itemize}
    \item \textbf{\textsc{FlippedPosData}}: We took the negative examples but misled the LLM by labeling them as if they were positive ("high-quality") examples.
    \item \textbf{\textsc{FlippedMixedData}}: We presented the LLM with the same set of positive and negative examples as MixedData, but swapped their labels (positives labeled as “low-quality” and negatives labeled as “high-quality”).
\end{itemize}

If the model adapted to the data, \textsc{FlippedPosData} should produce features for low-quality texts, and \textsc{FlippedMixedData} should invert the original \textsc{MixedData} features. Instead, we observed the opposite. As shown in Table \ref{expert_vs_baselines}, \textsc{FlippedPosData} produced features nearly identical to \textsc{PosData}, and \textsc{FlippedMixedData} closely matched \textsc{MixedData}. This suggests that flipping labels had minimal impact, indicating the model relied on prior notions of “high-quality” text rather than adapting to the input data.

\paragraph{(b) Is Pre-existing Knowledge the Primary Source of Generated Features?}
The above result raises the question: would the LLM have generated a similar list of features even if we gave it no data at all? To test this, we prompted the LLM to list features of high-quality text without providing any example data. Here, the model must rely solely on its internal knowledge.

\begin{itemize}
    \item \textbf{\textsc{NoData}}: The LLM is asked to imagine or recall what characteristics define good content, without seeing any specific dataset samples.
\end{itemize}

Remarkably, \textsc{NoData} features showed high recall of expert-defined criteria, strongly overlapping with \textsc{PosData} and \textsc{MixedData} (Table \ref{expert_vs_baselines}). Even without input data, the model reproduced nearly all expert criteria—fully for slogans and almost entirely for titles. This suggests the LLM relied on prior knowledge rather than capturing from the provided data, highlighting both its strong knowledge base and limited data-driven adaptation. While this demonstrates the model’s impressive knowledge base, it also underscores the lack of true data grounding in these direct generation approaches.


\begin{figure}[t]
  \centering
  \begin{subfigure}{\columnwidth}
  \includegraphics[width=\textwidth]{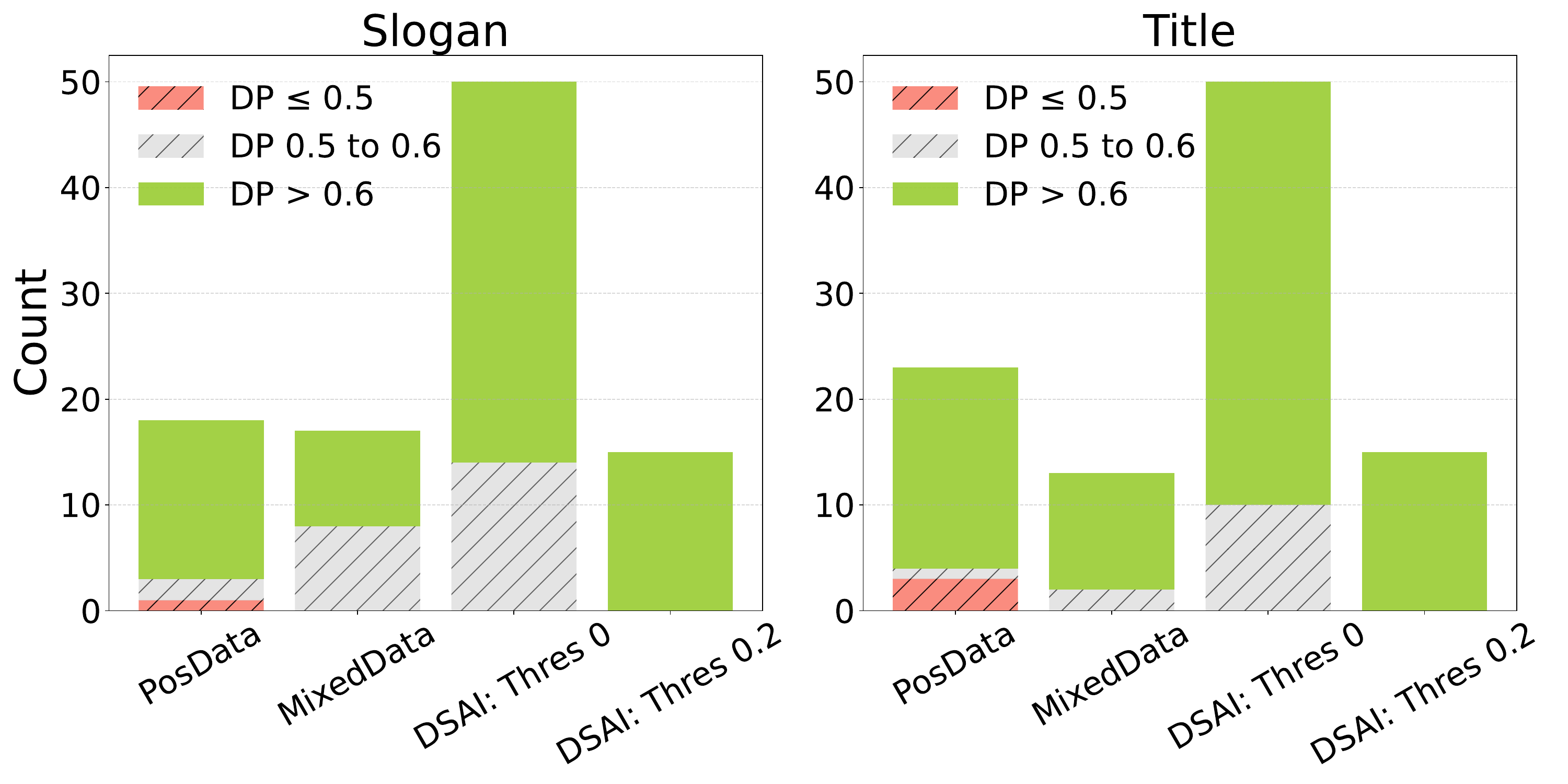}
  \caption{DSAI shows superior grounding over direct feature generation methods, with no DP score below 0.5. When prominence exceeds 0.2, all scores remain above 0.6.}
  \label{fig:llm_precision}
  \end{subfigure} \\
  \begin{subfigure}{\columnwidth}
    \includegraphics[width=\textwidth]{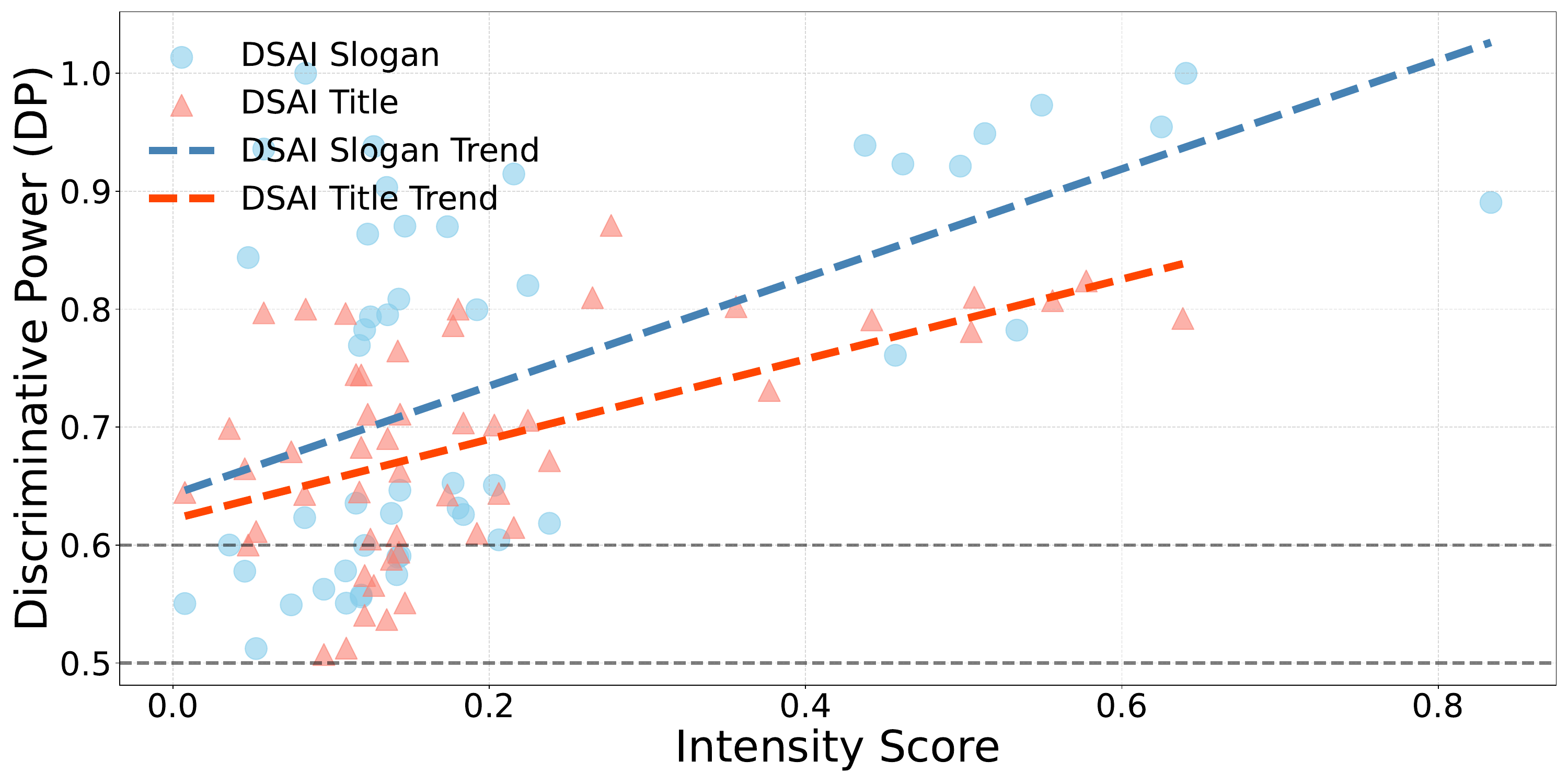}
  \caption{The graph shows that higher prominence leads to higher DP scores for DSAI features.}
  \label{fig:dsai_precision}
  \end{subfigure}
  \caption{DP scores for direct feature generation and DSAI methods.}
  \vspace{-3mm}
  
\end{figure}

\paragraph{(c) Are the Generated Features Truly Reflective of Positive Data's Latent Characteristics?}
We next examined the quality of the features generated by the LLM in terms of how well they actually characterize the positive class in the data. Just because a feature sounds like a good guideline doesn’t guarantee that it differentiates our positive and negative examples. For instance, the model might output “Use simple language” as a feature of good text, but if both our positive and negative examples equally exhibit (or fail to exhibit) this trait, then that feature isn’t really capturing what makes the positive group unique in our dataset.

To assess this, we quantify the discriminative power (DP) of each feature. We define a feature’s \textit{DP score} as a measure of how well that feature separates positive examples from negative ones in the dataset. In practice, we calculate DP score as the fraction of examples that exhibit the feature which belong to the positive class: $P(\textrm{positive}|\textrm{feature-present})$. One can think of this like a precision of the feature for identifying positive samples. A DP of 0.5 means the feature appears just as often in negatives as in positives – effectively no discriminative value. A DP closer to 1 means the feature is mostly present in positives (strong positive indicator), whereas a DP below 0.5 means the feature is actually more common in negatives, which would indicate a misidentified or inversely correlated feature.

Using this metric, we evaluated the features generated under the \textsc{PosData} and \textsc{MixedData} prompts. We found that several of those features had low DP scores, some even below 0.5 (Figure \ref{fig:llm_precision}). This means the model sometimes proposed features that were more prevalent in the “bad” examples than the “good” ones. These results highlight a limitation of directly using an LLM for feature generation: many of the features it outputs, even if they sound plausible, do not truly reflect the distinguishing characteristics of the positive data.



\paragraph{(d) Does Removing Task Context Improve Data Grounding?}
One hypothesis to improve grounding was that the LLM’s knowledge might be overly triggered by the context of the task in the prompt.
To test if removing task-specific context reduces bias and improves data grounding, we used:

\begin{itemize}
    \item \textbf{\textsc{NoContext}}: The LLM is given two unlabeled sets of texts but is not told which set is “high-quality” or “low-quality”, nor even that the goal is to identify high-quality text features. Essentially, we ask the model to compare two groups of texts without naming the task.
\end{itemize}

In this setting, the LLM defaulted to generating vague and generic descriptors (Appendix \ref{appendix:baseline6}). This suggests that task context removal alone does not improve data grounding and highlights the need for structured guidance during feature extraction.


\begin{figure*}[t]
\centering
  \includegraphics[width=0.95\textwidth]{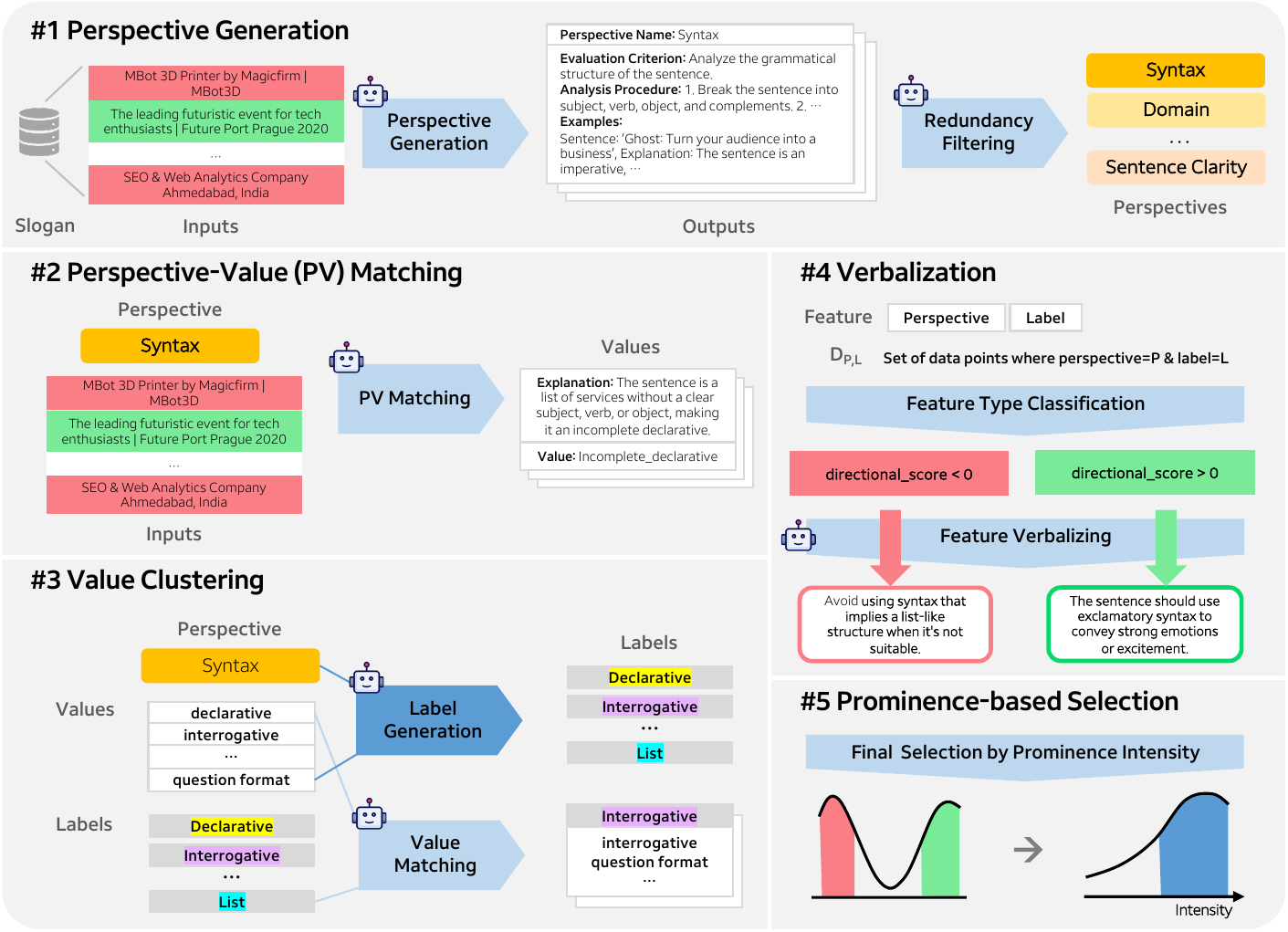}
  \caption{Overview of the DSAI pipeline: Perspectives are first generated to guide analysis (\#1), then used to match values to data points (\#2). These values are clustered to reduce redundancy (\#3), verbalized into concise criteria (\#4), and prioritized based on their prominence (\#5).}
  \label{fig:pipeline}
\end{figure*}

\section{Data Scientist AI}

DSAI is a five-stage framework for automated latent feature generation that leverages LLM capabilities within a structured process. The pipeline, illustrated in Figure \ref{fig:pipeline}, is designed to overcome the shortcomings identified in \S\ref{llm_data_grounding_issue} by guiding the LLM through controlled steps.


\paragraph{\#1 Perspective Generation}

In this first stage, we prompt the LLM to generate a diverse set of perspectives – these are different angles or aspects under which the data can be described. The key idea is to break down the analysis into multiple facets. To achieve this, we feed the LLM a small subset of the dataset, including a few positive and negative examples. By keeping the task context hidden, we minimize domain bias while still providing concrete data for the model to analyze. We then ask the LLM to propose distinct perspectives that might explain differences in the data. 

Each perspective generated by the LLM comes with a structured description: we instruct the model to give each perspective a name, a brief evaluation criterion, a suggested process for analyzing a text from that perspective, and an example from the provided data illustrating the perspective in action (Figure \ref{fig:pipeline} (1)). By having the LLM produce this structured output, we ensure consistency in how it conceptualizes each aspect. After generation, we also apply a de-duplication step to remove redundant or overlapping perspectives. The result of Stage 1 is a list of candidate perspectives, which serve as the conceptual foundation for the subsequent analysis.


\paragraph{\#2 Perspective-Value Matching}

In Stage 2, we systematically evaluate each data point against the perspectives identified in Stage 1. For every \textit{(perspective, data point)} pair, the LLM is prompted to assign a value that describes the data point with respect to that perspective. After this stage, each data point is associated with a set of \textit{(perspective, value)} pairs – one for each perspective – capturing how the LLM perceives that point on each aspect. 


\paragraph{\#3 Value Clustering}
To make the features more interpretable and reduce redundancy, we cluster similar values within each perspective.  We employ the LLM for this clustering task, drawing on its semantic understanding. The process works in two sub-steps for each perspective:
(1) Cluster Label Generation: The LLM examines all the values it assigned under a given perspective and proposes a smaller set of representative labels (cluster names).  (2) Value Assignment: Next, the LLM assigns each raw value to one of the generated cluster labels.

The outcome is that for each perspective, we now have a handful of feature categories (the cluster labels), each representing a group of similar values. Each data point can thus be described in terms of these cluster labels.


\paragraph{\#4 Verbalization}  

This stage transforms \textit{(perspective, label)} pairs into verbal \textit{criteria}. For each pair, we compute \( P(\textrm{positive} | D_{p,l}) \), the proportion of positive examples in dataset \(D_{p,l}\) corresponding to the \textit{(perspective, label)} pair $(p,l)$.  Using this, we calculate a \textit{directional score} \( 2 \times P(\textrm{positive}|D_{p,l}) - 1 \), which determines how the pair should be verbalized. Pairs with positive directional scores (>0) are directly verbalized as features describing positive data. Pairs with negative directional scores (<0) are transformed into "avoid" statements, which indirectly characterize positive data by specifying features to avoid.\footnote{For instance, if \textit{(clarity, low)} receives a negative directional score, it is verbalized as "Avoid sentences with low clarity."} This dual transformation approach ensures coverage of both desired and undesired traits.

\paragraph{\#5 Prominence-based Selection}  

Finally, DSAI employs \textit{prominence intensity} as the feature selection metric, defined as the absolute value of directional score $\|2*P(\textrm{positive}|D_{p,l}) - 1\|$. Using prominence as a metric, we can now select the most impactful features by setting a prominence threshold (which can be adjusted by the user) and retain only features above that threshold.
The key is that DSAI is not just outputting an unstructured list of features – it provides a way to prioritize them. This addresses the issue with the baseline LLM approach (\S\ref{llm_data_grounding_issue}), which gave no indication of which features were more important or reliable. By looking at the prominence scores, users can decide how many features to consider or where to draw the line between major and minor features.

After Stage 5, the final output of DSAI is a curated set of features, each in a clear natural-language form, typically accompanied by their prominence scores. These features are intended to be data-grounded and interpretable, providing insight into the data. In the next section, we evaluate how well this pipeline works in practice, especially in comparison to the direct LLM approach.



\section{Validation of Methodology Using Expert-Driven Annotation Dataset}

This section validates our methodology through experiments on various datasets, focusing on three key aspects: recall of expert-defined criteria (\S\ref{recall}), discriminative power of generated criteria (\S\ref{precision}), and reliability of pipeline modules (\S\ref{consistency_test}).

\begin{table*}[t]
\centering
\resizebox{\textwidth}{!}{
\begin{tabular}{l|r|r}
\Xhline{1.5\arrayrulewidth}
\textbf{Requirement} & \textbf{Prominence} & \textbf{Frequency} \\
\Xhline{1.5\arrayrulewidth}
\multicolumn{3}{c}{\textbf{Top Requirements}} \\
\Xhline{1.5\arrayrulewidth}
The advertising tone should convey a focus on quality, using optimistic and aspirational language. & 0.8571 & 14 \\
The sentence should convey an optimistic advertising tone that encourages engagement. & 0.8333 & 108 \\
The sentence should incorporate cultural references that align with consumptive themes. & 0.8333 & 12 \\
The sentence should employ indirect methods to engage the audience effectively. & 0.7857 & 28 \\
Ensure the sentence contains a component that emotionally appeals to the reader. & 0.7831 & 83 \\
\Xhline{1.5\arrayrulewidth}
\multicolumn{3}{c}{\textbf{Bottom Requirements}} \\
\Xhline{1.5\arrayrulewidth}
The sentence should fully and effectively communicate its intended message. & 0.0267 & 1,159 \\
The sentence should avoid merely providing information without an intended action or emotion. & 0.0250 & 1,122 \\
Ensure the sentence includes references to cultural significance. & 0.0248 & 1,009 \\
Avoid using imperative or overly complex grammatical structures in titles. & 0.0244 & 41 \\
Ensure the use of inclusive language in the sentence. & 0.0225 & 1,109 \\
Avoid sentences that lack necessary cultural references. & 0.0224 & 1,115 \\
\Xhline{1.5\arrayrulewidth}
\end{tabular}}
\caption{Top and Bottom Requirements of Slogan Dataset based on Prominence.}
\label{tab:slogan_top_bottom_requirements}
\end{table*}

\subsection{Recall of Expert-Defined Criteria} \label{recall}

One way to gauge DSAI’s effectiveness is to see if it can rediscover the ground-truth criteria that domain experts have defined for these datasets.
We applied the full DSAI pipeline to the slogans and research titles data annotated as described in \S\ref{llm-driven-data-analysis}.

\vspace{-1mm}
\paragraph{Criteria Generation} We generated criteria through our pipeline and retained those with $|D_{p,l}| >6$ and positive prominence intensity scores. This yielded 235 criteria for slogans and 198 criteria for research titles. Examples of the generated criteria are shown in Table \ref{tab:slogan_top_bottom_requirements}.

\vspace{-1mm}
\paragraph{Human Feature Matching} For recall evaluation, one annotator initially performed loose matching of generated criteria, which was then validated through majority voting among three annotators.

\paragraph{Recall}
Our methodology showed strong performance in reproducing expert-defined criteria, even at high prominence intensity thresholds. All 9 expert criteria for slogans were captured at a threshold of 0.348, and 83\% recall (10/12) was achieved for research titles at a threshold as high as 0.692 (Table \ref{expert_vs_baselines}). While \textsc{PosData} and \textsc{MixedData} also showed decent coverage, their results relied on LLM's pre-existing knowledge as discussed in \S\ref{llm_data_grounding_issue}. In contrast, our approach, by design, minimizes such potential bias by withholding task-specific context, while still achieving comparable or better recall rates.

\vspace{-1mm}
\paragraph{Recall Dynamics across Various Thresholds} The adjustable prominence intensity thresholds allow users to tailor their analyses by balancing discriminative power and coverage. We provide a detailed analysis of recall dynamics across different threshold values in Appendix \ref{appendix:threshold_analysis}.
In summary, we categorized expert-defined criteria, ranked their importance, and examined at which thresholds they were filtered out. More important criteria persisted at higher thresholds, while less critical ones were eliminated at lower thresholds.


\vspace{-1mm}
\subsection{Discriminative Power (DP)}
\label{precision}

Having shown that DSAI can reproduce known criteria, we next investigate whether the prominence score we assign to features actually correlates with real discriminative power. Intuitively, if our pipeline is working correctly, features with higher prominence intensity should be the ones that better distinguish positive from negative examples.

In our evaluation, we took all the DSAI-generated criteria and binned them into five buckets based on prominence intensity scores. From each bucket, we sampled 10 criteria for manual examination. 
For each sampled feature, we went back to the dataset and annotated instances to see if the feature was present or absent  in those examples. We excluded non-applicable\footnote{\textit{e.g.}, feature about “use of sales terms” in examples unrelated to business models.} cases from the calculations.

Using these annotations, we computed the \textit{DP score} of each feature as mentioned earlier (\S\ref{llm_data_grounding_issue}(c)), and then examined the DP scores across the different prominence buckets.

{\small
\begin{align*} \label{eqn:my_equation}
DP\text{ }Score = \left\{
\begin{aligned}
    &P(\textrm{positive}|\textrm{feature-present}) \quad \text{if } \\
    &\quad \quad \quad \quad  \quad \textrm{directional\_score} > 0, \\
    &P(\textrm{negative}|\textrm{feature-absent}) \quad \text{elif } \\
    &\quad \quad \quad \quad \quad  \textrm{directional\_score} < 0.
\end{aligned}
\right.
\end{align*}
}%

The results confirmed our expectations: criteria with higher prominence scores generally showed higher DP. As illustrated in Figure \ref{fig:dsai_precision}, all of the DSAI-generated criteria achieved DP > 0.5. Lower-prominence features, in contrast, were sometimes borderline 0.5, reinforcing the idea that prominence is a good indicator of a feature’s reliability. In short, by using the prominence intensity metric, DSAI effectively filters and ranks features by their true discriminative power. This validation not only demonstrates the reliability of our pipeline’s outputs, but also highlights that the prominence metric can guide users to the most trustworthy features. Practically, this means one can focus on high-prominence features for critical decisions, knowing they have been quantitatively vetted to distinguish positive examples well.



\vspace{-1mm}
\subsection{Reliability of Pipeline Operations}
\label{consistency_test}

DSAI’s multi-stage pipeline relies on the LLM’s output at several steps. 
While the previous sections show the end results are effective, we also wanted to ensure that each intermediate step was performed reliably by the LLM. To assess this, we had the LLM perform a self-check on its work for certain stages. Each verification was conducted in a separate session from the original task to ensure independence. We focused on the stages that have a well-defined objective where consistency can be measured:
\begin{itemize}
    \item Stage 2 (Value Matching): For each perspective and data point, after the LLM assigned a value, we asked the LLM to confirm whether that assignment was correct given the data point’s content. 
    \item Stage 3 (Clustering): After the LLM clustered values, we gave it each value along with the cluster label and asked if that assignment was appropriate.
    \item Stage 4 (Verbalization): We asked the LLM to verify that each verbalized feature correctly described the intended cluster and perspective, and that the phrasing (direct or “avoid”) corresponded to the sign of the directional score.
\end{itemize}

The LLM’s self-audit verification process showed high consistency rates of >98\%, \>94\%, and 98\% for stages \#2, \#3, and \#4 respectively. These results indicate that the pipeline’s internal operations are reliable; the LLM is largely consistent and does not contradict itself when asked to re-check its work.

\vspace{-1mm}
\section{Real-World Application with Quantitative Datasets}
\vspace{-1mm}

\begin{figure*}[t]
  \centering
\includegraphics[width=\textwidth]{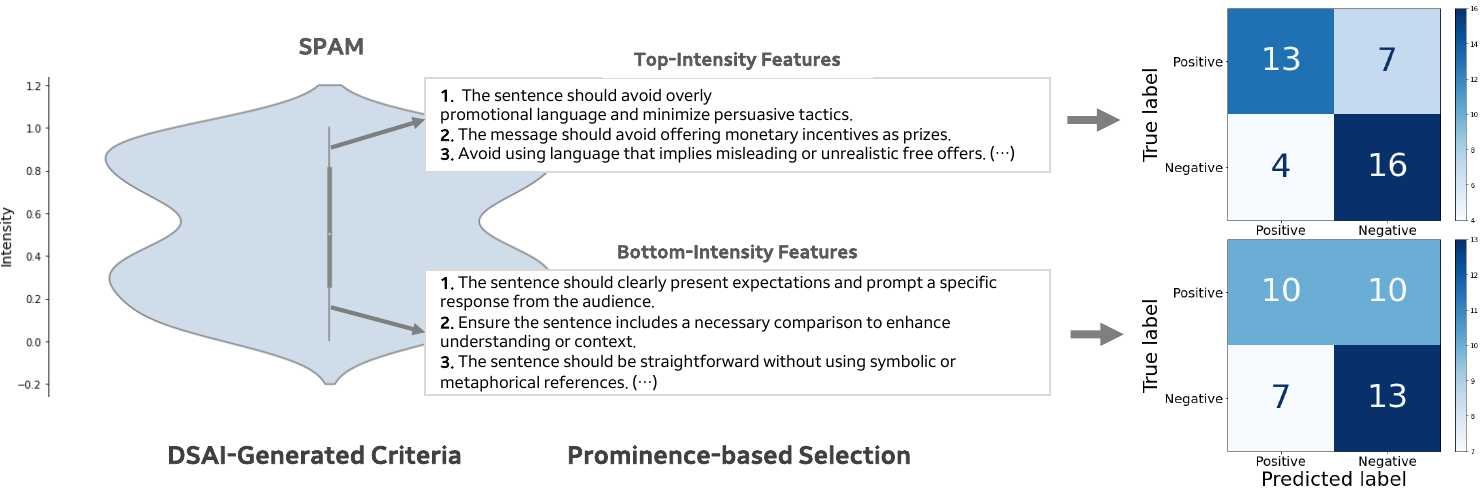}
  \caption{Example of interpretable spam classification: The figure shows how feature prominence guides criteria selection, with high-prominence criteria improving spam classification performance.}
  \label{fig:spam_cls}
\end{figure*}

Having validated DSAI on datasets with known ground truth criteria, we now apply it to several real-world datasets to demonstrate its practicality and versatility. We selected three real-world user feedback datasets critical for business insights \cite{luo2022empirical}: (1) MIND \cite{dataset_mind}, analyzing engagement features in news headlines with high CTR as the positive group; (2) spam detection \cite{dataset_spam}, identifying patterns in spam messages as the positive group; and (3) Reddit \cite{dataset_reddit}, exploring interaction-promoting linguistic features in highly upvoted comments as the positive group.
These datasets differ significantly in content and domain, which allows us to see how DSAI adapts to different domains without any domain-specific tuning.

\vspace{-1mm}
\paragraph{Prominence Distribution and Sample Insights} 

Running DSAI on each dataset, we observed that the distribution of feature prominence scores differs by domain. This is expected: each domain has its own characteristics and noise levels, so the threshold for what constitutes a strongly discriminative feature will vary. For instance, in the spam dataset, it might be crucial to set a higher prominence threshold to avoid any features that could lead to false positives. In the news headline dataset, one might choose a threshold that balances identifying strong engagement drivers while not missing out on subtler but interesting patterns. The Reddit data might show a different spread, capturing nuances of informal language or humor that drive upvotes.

Importantly, DSAI captured not only general traits (like "uses urgent language" might be a spam trait common across many messages, or "mentions specific names/events" for news headlines) but also fine-grained nuances specific to subsets of each dataset (like "sarcastic undertone" or "emotionally intense"). These are the kinds of details often overlooked by simpler LLM analyses or manual inspection. For example, DSAI might find that in the news dataset, a certain style of phrasing has a subtle impact on CTR, which wouldn’t be obvious without this kind of analysis. These observations highlight the benefit of a data-centric approach: DSAI can adapt to the particular domain and context of each dataset, rather than relying on one-size-fits-all features. We provide detailed breakdowns and examples from each dataset in Appendix \ref{appendix:Industry_generated_features}.




\vspace{-1mm}
\paragraph{Potential for Downstream Tasks}

Because DSAI produces human-readable criteria, the extracted features have the potential to readily support various downstream applications such as style transfer (rewriting content to meet certain criteria), generating annotation guidelines (for human labelers to follow), or directly for classification tasks.
 
To illustrate this, we conducted a toy spam classification experiment using 20 spam and 20 ham samples. Using the five criteria with the highest prominence intensity scores led to effective classification performance, while using the five criteria with the lowest prominence resulted in poor performance (See Figure \ref{fig:spam_cls}). This small experiment demonstrates that DSAI’s prominence scoring correlates with real utility: the features deemed important by DSAI indeed helped in a classification task, while those deemed unimportant were not useful.

\vspace{-1mm}
\section{Conclusion}
\vspace{-1mm}
In this paper, we proposed DSAI, a faithful data-driven feature extraction framework that ensures LLMs identify latent characteristics from data without relying on their domain-related biases. DSAI automates thorough examination of large datasets while minimizing human labor and enhancing interpretability through source-to-feature traceability. Through empirical validation, we confirmed its capability to extract meaningful features, suggesting its potential for applications requiring interpretable and efficient feature extraction.


\section{Limitation}

While our results confirm that DSAI can effectively guide LLMs to produce data-grounded features, there are still several limitations to consider. First, the framework’s performance depends heavily on the quality of the underlying LLM. If the model struggles to assign values, cluster them appropriately, or generate coherent perspectives, the outputs can be error-prone or may require significant manual intervention. Second, although GPT-4o was used for annotation and evaluation in some of our experiments, relying on model-based annotations can introduce bias or label noise, potentially affecting overall accuracy. We mitigated this risk by referencing expert-defined criteria and performing manual reviews, but model-based annotations remain a potential source of error.

In addition, DSAI’s current design has thus far been demonstrated primarily on text data, so extending it to other modalities like images, audio, or structured logs may require specialized adaptations or the use of different LLMs. Another limitation involves computational cost and scalability: when evaluating data points across multiple perspectives and then clustering them, running DSAI on very large datasets could become computationally expensive. Employing optimization or sampling strategies might therefore be necessary to maintain efficiency. Finally, despite reducing the reliance on domain experts, some degree of human oversight may still be necessary, particularly in high-stakes environments such as legal or medical settings where interpretability and correctness are paramount.



\bibliography{custom}

\newpage

\appendix

\section{LLM Annotation with Expert-Defined Criteria}
\label{appendix:llm_annotation}

This section describes our methodology employed for data annotation using expert-defined criteria and evaluates the effectiveness of each criterion through distribution analysis.

\subsection{Method}

We established a gold standard by manually annotating 10–20 samples per dataset and using expert-defined criteria. After optimizing prompts to maximize alignment with this manual annotation, we used the most aligned prompt to annotate 3,000 data points with GPT-4o, classifying each as "Feature-Present" or "Feature-Absent". This score for each data point was calculated as the total number of conforming criteria. Based on these scores, we selected the top and bottom 600 samples as \textit{positive} (high-quality text) and \textit{negative} (low-quality text) samples respectively.

This annotation process proved cost-effective compared to human annotation, with an average cost of \$3.5 USD per 3,000 annotations.

\begin{table*}[t]
\centering
\resizebox{0.7\textwidth}{!}{
\begin{tabular}{l|r|r}
\Xhline{0.4mm}
\textbf{Criterion} & \textbf{Top Feature-Present (\%)} & \textbf{Bottom Feature-Present (\%)} \\ 
\hline \hline
Direct/Straight-forward & 99 & 61 \\
Consise but not too simple & 100 & 0 \\ 
Pleasant to hear & 84 & 1 \\ 
Includes a sales idea & 98 & 4 \\ 
No exaggeration & 99 & 91 \\ 
Future-oriented & 81 & 1 \\ 
Clear positioning & 93 & 2 \\ 
Highlight the brand's unique traits & 99 & 0 \\ 
Include the brand name & 99 & 30 \\ 
\Xhline{0.4mm}
\end{tabular}}
\caption{Ratio of Feature-Present in Top and Bottom for Slogan expert-defined criteria.}
\label{tab:slogan_top_bottom}
\end{table*}

\subsection{Ensuring Distinction Between Positive and Negative Groups}

Given that some expert criteria may lack sufficient discriminative power despite their theoretical importance, we conducted an analysis comparing the ratios of "Feature-Present" for each criterion between top- and bottom-ranked samples. Note that a truly discriminative criterion should show high Feature-Present ratio in the top group and low Feature-Present ratio in the bottom group.

\begin{table*}[t]
\centering
\resizebox{0.7\textwidth}{!}{
\begin{tabular}{l|r|r}
\Xhline{0.4mm}
\textbf{Checklist Item} & \textbf{Top Feature-Present (\%)} & \textbf{Bottom Feature-Present (\%)} \\ \hline \hline
Simple format & 100 & 3 \\ 
Direct & 100 & 1 \\ 
Informative and specific & 100 & 0 \\ 
Functional (with scientific keywords) & 100 & 3 \\ 
Concise and precise & 100 & 0 \\ 
Include the main theme & 100 & 0 \\ 
Not too long or too short & 100 & 3 \\ 
Avoid whimsical words & 100 & 60 \\ 
Avoid jargon & 100 & 44 \\ 
Mention place/sample size if valuable & 100 & 72 \\ 
Important terms at the beginning & 100 & 11 \\ 
Descriptive titles preferred & 100 & 9 \\ 
\Xhline{0.4mm}
\end{tabular}}
\caption{Ratio of Feature-Present in Top and Bottom for Title expert-defined criteria.}
\label{tab:title_top_bottom}
\end{table*}

\subparagraph{Slogan (Top vs. Bottom)}
Table \ref{tab:slogan_top_bottom} demonstrate significant Feature-Present ratio disparities between the top and bottom groups for most criteria. Several criteria demonstrate maximal distinction (\textit{e.g.}, "Concise but not too simple" (100\% vs. 0\%)), and some show moderate but still meaningful differentiation (\textit{e.g.}, "Include the brand name" (99\% vs. 30\%)). "No exaggeration" (99\% vs. 91\%) and "Direct/Straight-forward" (99\% vs. 61\%) have smaller gaps.

\subparagraph{Title (Top vs. Bottom)} Table \ref{tab:title_top_bottom} shows even stronger distinction between the top and bottom groups. Multiple criteria (\textit{e.g.}, "Simple format" "Direct", "Concise and precise") achieve 100\% Feature-Present ratio in the top group compared to as low as 0–3\% Feature-Present ratio in the bottom group, indicating their high predictive value for title quality. The large differences across all criteria confirm their strong discriminative power.

\section{Implementation Details and Cost Analysis}

\paragraph{Details of Each Inference} In Stage \#1 Perspective Generation, we generated 50 perspectives at per forwarding step across three steps, iteratively concatenating each step's output with the few-shot example in the prompt for the next step. For Stage \#2 Perspective-Value Matching, each forwarding step processed one sentence and three perspectives as input. Stage \#3 Perspective-Oriented Value Clustering used all generated values per perspective in order to effectively cluster shared characteristics. In Stage \#4 Verbalization, we transformed each perspective-label pair into a compact sentence (\textit{criterion}). The prompts used in this process will be released through github repository.

\paragraph{Cost Analysis} 

The total cost of processing 10 perspectives and 100 sentences is \$2.43746, broken down as follows:

In \textbf{Step 1: Perspective Generation}, generating 10 perspectives costs \$0.0304. Each perspective costs \$0.00304. With a total of 2,100 input tokens and an average of 82.90 output tokens per perspective, the cost is calculated as:
\[
10 \times 0.00304 = 0.0304
\]

In \textbf{Step 2: Perspective-Value Matching}, each sentence requires \( \lceil 10/3 \rceil = 4 \) inferences, as each inference processes 3 perspectives. For 100 sentences, this results in:
\[
100 \times 4 = 400 \text{ inferences.}
\]
With a cost per inference of \$0.00496, the total cost for this step is:
\[
400 \times 0.00496 = 1.984
\]
On average, each inference involves 2,206 input tokens and 440 output tokens.

In \textbf{Step 3: Perspective Value Clustering}, each perspective incurs a clustering cost of \$0.0087125. This includes two components:
\begin{itemize}
    \item \textbf{Label Generation}, costing \$0.00579625 for 597 input tokens and approximately 1,010 output tokens.
    \item \textbf{Value Matching}, costing \$0.00291625 for 269 input tokens and 516 output tokens.
\end{itemize}
The total clustering cost per perspective is:
\[
0.00579625 + 0.00291625 = 0.0087125
\]
For 10 perspectives, the total clustering cost is:
\[
10 \times 0.0087125 = 0.087125
\]

In \textbf{Step 4: Verbalization}, it is assumed that the 10 perspectives cluster into 5 groups each, resulting in 50 perspective-label pairs. Verbalizing each pair costs \$0.0018, so the total cost for this step is:
\[
50 \times 0.0018 = 0.09
\]

Summing all the costs, the total processing cost is:
\[
0.0304 + 1.984 + 0.087125 + 0.09 = 2.43746
\]

\section{Prominence Intensity and Occurrence Analysis}

This section explores the relationship between feature frequency, prominence, and their impact on discriminative power.
By analyzing feature occurrence and prominence distributions, we provide insights to help readers determine appropriate thresholds for these factors, aiding in the selection of features that optimize performance in distinguishing positive and negative traits within datasets.

\begin{figure}[t]
  \centering
  \begin{subfigure}{0.48\textwidth}
    \includegraphics[width=\textwidth]
    {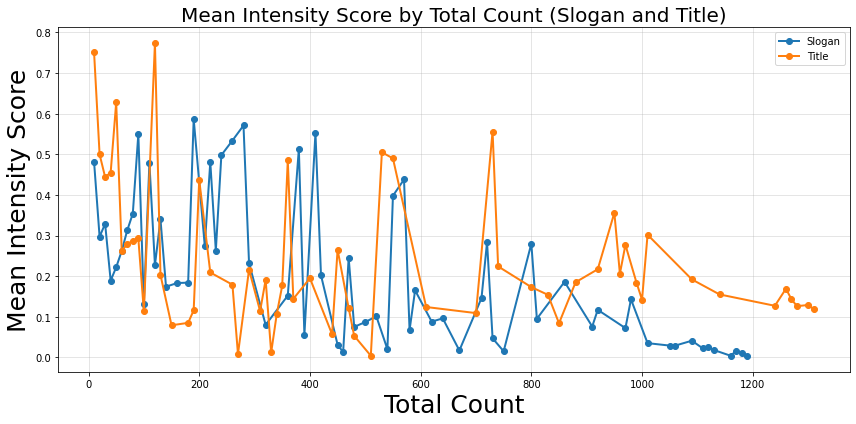}
    \caption{Average of Prominence Score per Frequency Bucket}
    \label{fig:mean_intensity_score_by_total_count}
  \end{subfigure}
  
  \begin{subfigure}{0.48\textwidth}
    \includegraphics[width=\textwidth]{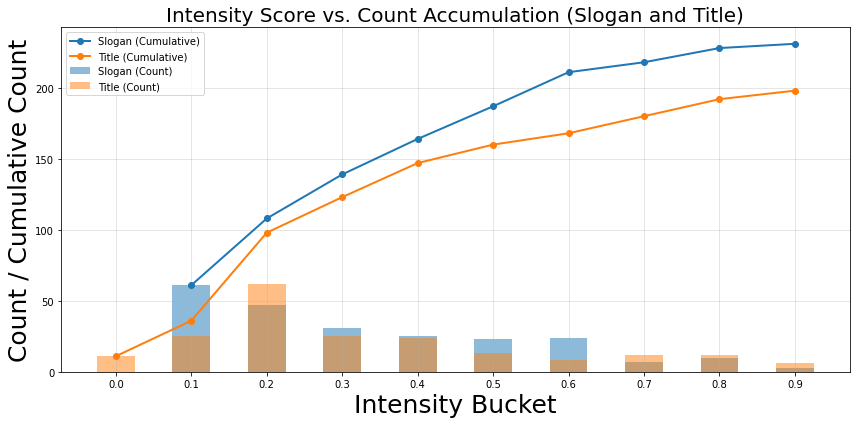}
        \caption{Data Coverage per Prominence Bucket}
    \label{fig:count_thres_intensity}
  \end{subfigure}
  \caption{Comparison of Prominence scores and data coverage across frequency and prominence buckets.}
  \label{intensity_count}
\end{figure}

\subsection{Influence of Feature Occurrence on Discriminative Power}

We analyzed the influence of feature occurrence frequency on discriminative power by grouping features into frequency buckets (e.g., <=10, 20-100, and >100) and evaluating mean/maximum prominence metrics (Figure \ref{fig:mean_intensity_score_by_total_count})

\paragraph{Low-Frequency Features} Features with low frequencies (<= 10) demonstrated higher discriminative power, with mean prominence as high as 0.482 (some reaching even 1.0) in the slogan dataset. This suggests that low-frequency features effectively represent specific traits of positive data, though their rarity may limit generalization.

\paragraph{Medium-Frequency Features} Features with medium frequencies 20–100) displayed a balance between specificity and generality. They demonstrated consistent performance in discriminative tasks through moderate mean prominence values.

\paragraph{High-Frequency Features} Features with high frequency (> 100) were found to be more generic, showing low mean prominence values (0.087 in the slogan dataset for frequencies > 500). Though less discriminative, they provide valuable insights into the dataset's baseline characteristics. Their high frequency makes them suitable for applications where specificity is less critical.

\subsection{Influence of Feature Prominence on Discriminative Power}

We also analyzed the distribution of features across different prominence score levels of by examining frequency accumulation within prominence buckets (Figure \ref{fig:count_thres_intensity}). This analysis revealed how features were concentrated across low, moderate, and high prominence scores.

\paragraph{Slogan} The features were concentrated in the lower prominence range, with 61 features below 0.1 and 108 features below 0.2. This distribution suggests that while the majority of features exhibit limited discriminative power, a select subset of features with higher intensities plays a disproportionate role in capturing latent positive traits.

\paragraph{Title} An even more pronounced skew toward lower intensities was observed, with 11 features showing zero prominence and 62 features below 0.2. This distribution highlights that while low-prominence features comprise the majority of the dataset, they have limited effectiveness in distinguishing positive traits.

\section{Threshold Analysis on Slogans}
\label{appendix:threshold_analysis}

In this section, we track how different prominence and frequency thresholds affect the recall of expert criteria, with particular attention to which criteria types are most resistant to threshold increases. 
In general, we observe progressive filtering of features as thresholds increase, which aligns with our criteria importance categorization. However, we also observed some exceptions diverging from their importance categorization. These findings illustrate how DSAI can identify discrepancies between theoretical feature importance and actual implementation patterns in the data.

\subsection{Analysis of Expert-defined Criteria}

We categorize expert criteria based on their importance to slogan effectiveness, ranging from critical to supplementary features. This categorization provides a framework for analyzing which features persist across different thresholds.

\paragraph{(a) Core Message Delivery features (Critical Importance)}
These features are essential for effective communication of the brand's core sales proposition and unique characteristics. They are considered critical because they directly influence a slogan's ability to capture and deliver the brand's main message to consumers. 

\begin{itemize}
    \item 4. Convey the sales idea clearly and concisely
    \item 8. Emphasize the brand's unique traits or the benefits it provides
\end{itemize}

\paragraph{(b) Structure and Expression features (High Importance)}
These features focus on clarity and readability, making a slogan easy to understand and leaving a lasting impression. They significantly impact audience engagement and retention, though they may not directly affect the message's content.

\begin{itemize}
    \item 1. It should be direct and straightforward
    \item 2. Keep it simple, but not overly simple
    \item 5. Avoid misleading or exaggerated words
\end{itemize}

\paragraph{(c) Tone and Atmosphere features (Moderate Importance)}
These features create an emotional connection with the audience through engaging and positive tone. While they enhance a slogan's appeal, they are not as pivotal as core message delivery or structural clarity.

\begin{itemize}
    \item 3. It should have a pleasant tone
    \item 6. It should be future-oriented
\end{itemize}

\paragraph{(d) Supplementary features (Lower Importance)}
These features can enhance the slogan's overall quality or effectiveness, but are not essential for a slogan's primary function.

\begin{itemize}
    \item 9. Include the brand name in the slogan
    \item 7. Clear positioning through comparison or closeness
\end{itemize}

\subsection{Prominence Intensity Threshold Analysis}

Table \ref{slogan_intensity} displays the effect of different prominence thresholds on recall and data size.
\begin{table}[h!]
    \centering
    \resizebox{\columnwidth}{!}{
    \begin{tabular}{r|r|r}
        \Xhline{0.4mm}
        \textbf{Recall} & \textbf{Threshold (Prominence)} & \textbf{Data Size} \\
        \hline \hline
        1 & 0.003 & 235 \\
        0.889 & 0.348 & 83 \\
        0.778 & 0.549 & 44 \\
        0.667 & 0.714 & 15 \\
        0.556 & 0.750 & 12 \\
        0.444 & 0.833 & 6 \\
        0.333 & 0.857 & 5 \\
        0 & 1 & 0 \\
        \Xhline{0.4mm}
    \end{tabular}}
    \caption{Recall and Data Size Across Prominence Thresholds}
    \label{slogan_intensity}
\end{table}

Recall remains stable until reaching a relatively high threshold (0.348), indicating robust baseline coverage of our pipeline. 
The relationship between importance levels and threshold resilience shows an overall aligned pattern, though with some complexity: while lower-importance criteria are consistently filtered out early, criteria of moderate importance and above show varied retention patterns, with some excluded at mid-level thresholds and others persisting until the highest thresholds.

\begin{figure*}[t]
  \centering
    \includegraphics[width=\textwidth]{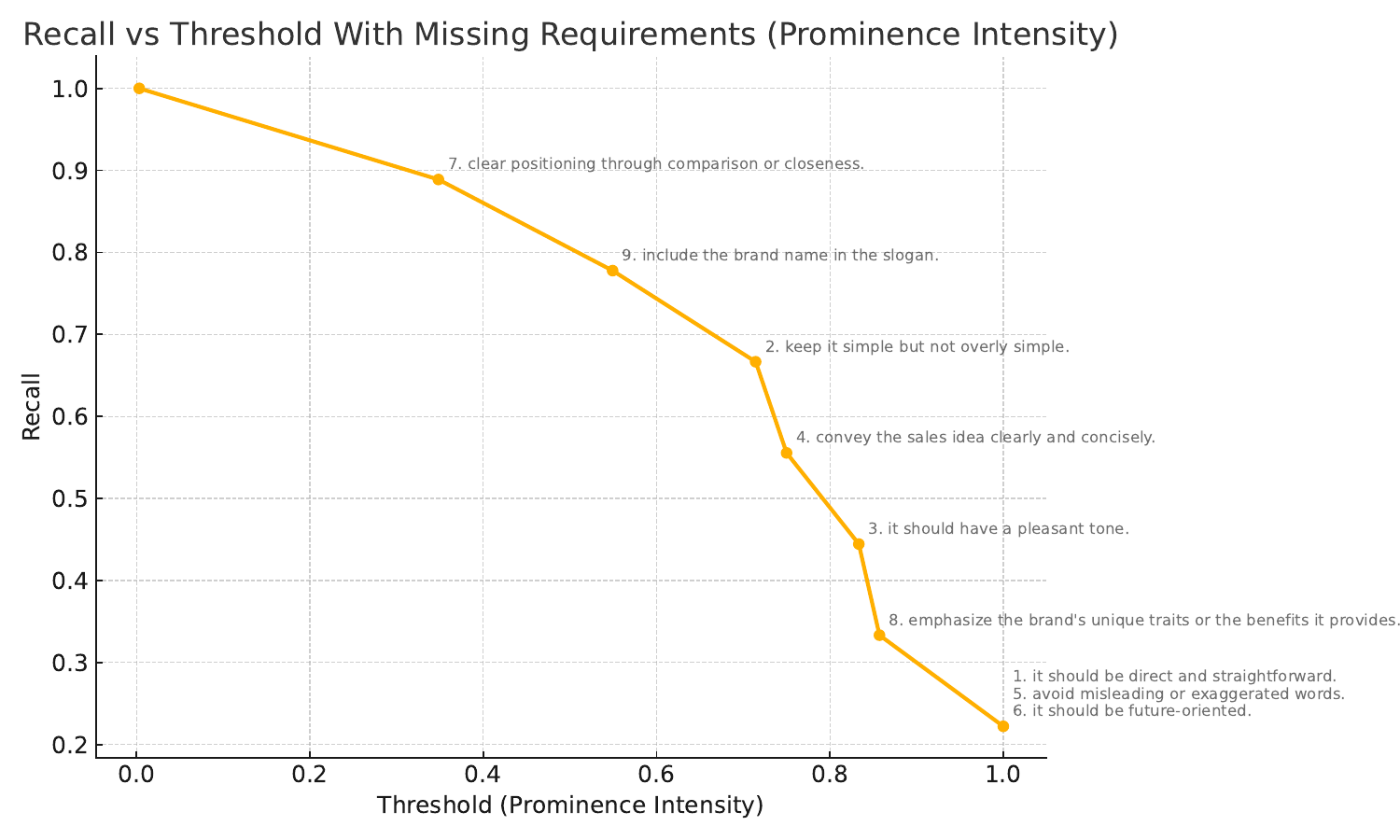}
    \caption{Dropped criterion as Prominence threshold increases}
    \label{fig:subfig1}
\end{figure*}

\begin{enumerate}
    \item \textbf{Early Exclusions (<0.6)}: \textit{7. Clear positioning through comparison or closeness} (lower importance) and \textit{9. Include the brand name in the slogan` (lower importance)} are excluded first, consistent with their supplementary nature.
    \item \textbf{Mid-level Exclusions (<0.8)}: General criteria such as \textit{2. Keep it simple but not overly simple} (high importance) and \textit{4. Convey the sales idea clearly and concisely} (critical importance) are excluded only at higher thresholds (0.721 and above), underscoring their broad applicability.
    \item \textbf{Late (No) Exclusions}: Multiple features with moderate (\textit{6. Future-oriented}) or above (\textit{e.g.}, \textit{1. Directness} (high importance), \textit{8. Emphasize the unique traits and benefits} (critical importance)) persist until the highest thresholds. 
\end{enumerate}


\subsection{Frequency Threshold Analysis}

Table \ref{tab:below1} illustrates how recall changes as we filter features based on their frequency in the dataset. 
\begin{table}[h!]
    \centering
    \resizebox{\columnwidth}{!}{
    \begin{tabular}{r|r|r}
        \Xhline{0.4mm}
        \textbf{Recall} & \textbf{Threshold (Frequency)} & \textbf{Data Size} \\
        \hline \hline
        1 & 5 & 236 \\
        0.889 & 93 & 86 \\
        0.778 & 217 & 66 \\
        0.667 & 571 & 38 \\
        0.556 & 661 & 31 \\
        0.444 & 1,005 & 16 \\
        0.333 & 1,115 & 9 \\
        0.111 & 1,192 & 2 \\
        0 & 1,196 & 1 \\
        \Xhline{0.4mm}
    \end{tabular}}
    \caption{Recall and Data Size Across Frequency Thresholds}
    \label{tab:below1}
\end{table}

Recall remains stable up to a frequency threshold of 93, indicating strong coverage of expert criteria even when considering only frequently observed patterns. 
The relationship between theoretical importance and threshold resilience reveals both expected alignments and interesting disparities. Critical features persist at high thresholds, and certain lower-importance features being filtered early. Some theoretically important features drop out earlier than expected, while certain lower-importance features demonstrate surprisingly high frequency in practice.


\begin{figure*}[h]
  \centering
    \includegraphics[width=\textwidth]{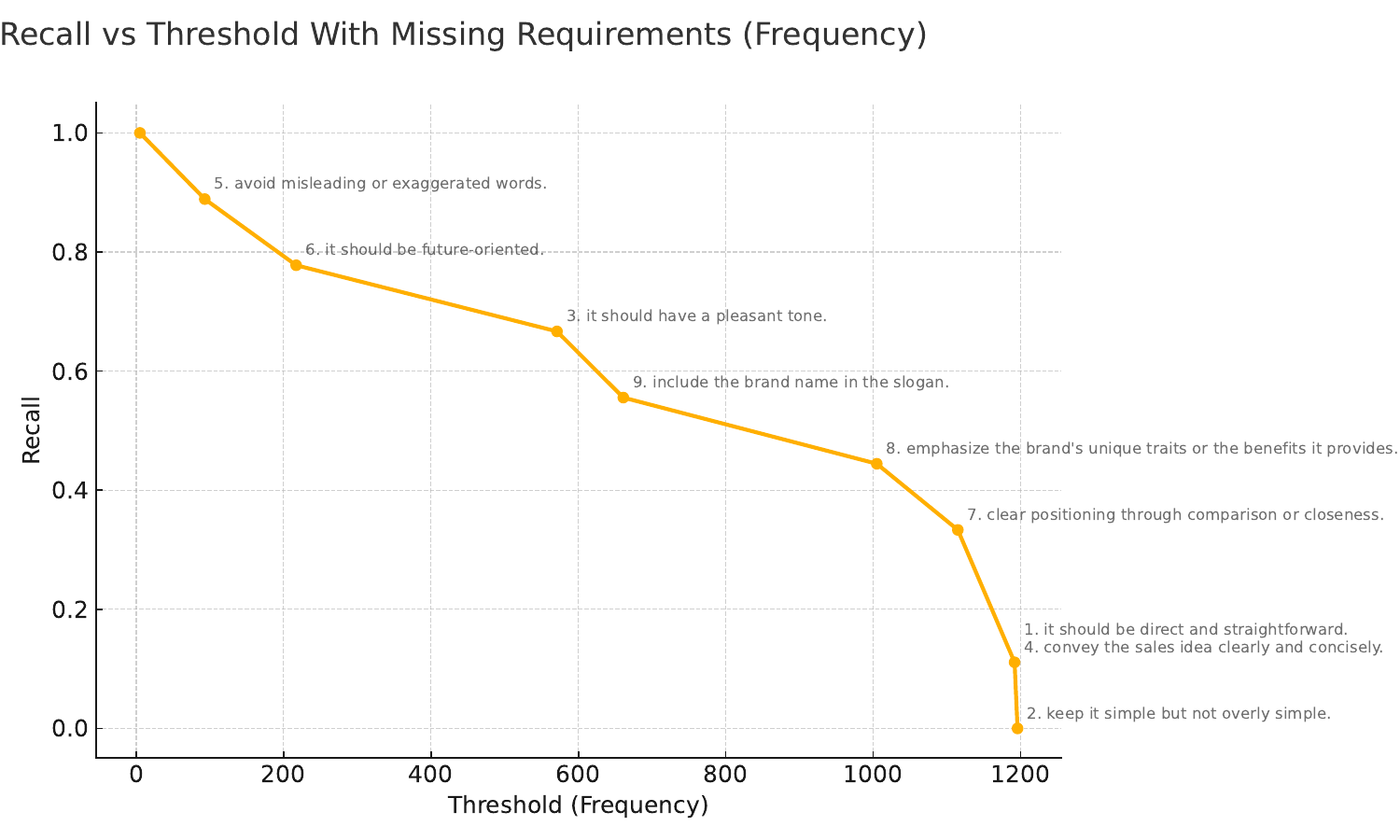}
    \caption{Dropped criterion as Frequency threshold increases.}
  \label{slogan_threshold}
\end{figure*}

\begin{enumerate}
    \item \textbf{Early Exclusions (<500)}: Notably, \textit{5. Avoid misleading words} (high importance) and \textit{6. Future-oriented tone} (moderate importance) drop out first. According to our theoretical categorization, these are not supplementary features, yet they appear less frequently in actual slogans than their theoretical significance would suggest. 
    \item \textbf{Mid-level Exclusions (<1000)}: \textit{3. Pleasant tone} (moderate importance) and \textit{9. Include brand name} (lower importance) are filtered out at moderate thresholds, aligning with their moderate importance categorization.
    \item \textbf{Late Exclusions}: Multiple core features like \textit{4. Convey the sales idea clearly} (critical importance) and \textit{1. Direct and straightforward} (high importance) has high frequency, confirming that these features are both theoretically critical and practically prevalent. 
\textit{7. Clear positioning through comparison} (lower importance) shows a notable deviation between theoretical and practical implementation.
\end{enumerate}

\section{Threshold Analysis on Research Titles}

Similar to our analysis on slogans (Appendix \ref{appendix:threshold_analysis}), the results suggest that DSAI can effectively capture the nuanced reality of title construction, where practical implementation patterns may differ from theoretical guidelines.

\subsection{Analysis of Expert-defined Criteria}
We begin by categorizing expert-defined criteria based on their contribution to a title's primary function: effective delivery of research content to readers.

\paragraph{Core Information Delivery features (Critical Importance)}
These features are fundamental as they directly affect a title's ability to enable readers quickly grasp the paper's topic and contributions.
\begin{itemize}
    \item 3. The title needs to be informative and specific
    \item 4. The title needs to be functional (with essential scientific "keywords") 
    \item 6. The title should include the main theme of the paper
\end{itemize}

\paragraph{Structural and Format features $\rightarrow$ Readability (High Importance)}
These features optimize the title's readability and clarity. While they do not directly affect content, they are crucial for successful information delivery.
\begin{itemize}
    \item 1. The title needs to be simple in terms of format
    \item 2. The title needs to be direct
    \item 5. The title should be concise and precise
    \item 7. The title should not be too long or too short
\end{itemize}

\paragraph{Linguistic Expression features (Moderate Importance)}
These features maintain academic professionalism while ensuring accessibility. They enhance the title's effectiveness without being critical to its basic function.
\begin{itemize}
    \item 8. The title should avoid whimsical or amusing words
    \item 9. The title should avoid non-standard abbreviations and unnecessary acronyms (or technical jargon)
\end{itemize}

\paragraph{Supplementary Guidelines (Lower Importance)}
These guidelines can be advantageous in specific contexts but are not universally essential.
\begin{itemize}
    \item 10. Place of the study and sample size should be mentioned only if it adds to the scientific value of the title
    \item 11. Important terms/keywords should be placed at the beginning of the title
    \item 12. Descriptive titles are preferred to declarative or interrogative titles
\end{itemize}

\subsection{Prominence Intensity Threshold Analysis}

\begin{table}[h!]
    \centering
    
    \resizebox{\columnwidth}{!}{
    \begin{tabular}{r|r|r}
        \Xhline{0.4mm}
        \textbf{Recall} & \textbf{Threshold (Prominence)} & \textbf{Data Size} \\
        \hline \hline
        0.833 & 0 & 199 \\
        0.750 & 0.692 & 40 \\
        0.667 & 0.778 & 29 \\
        0 & 1 & 0 \\
        \Xhline{0.4mm}
        
    \end{tabular}}
    
    \caption{Recall and Data Size Across Prominence Thresholds}
    \label{tab:below3}
\end{table}

Table \ref{tab:below3} demonstrates the impact of increasing the prominence threshold on the recall and data size.

\begin{figure*}[t]
  \centering
    \includegraphics[width=\textwidth]{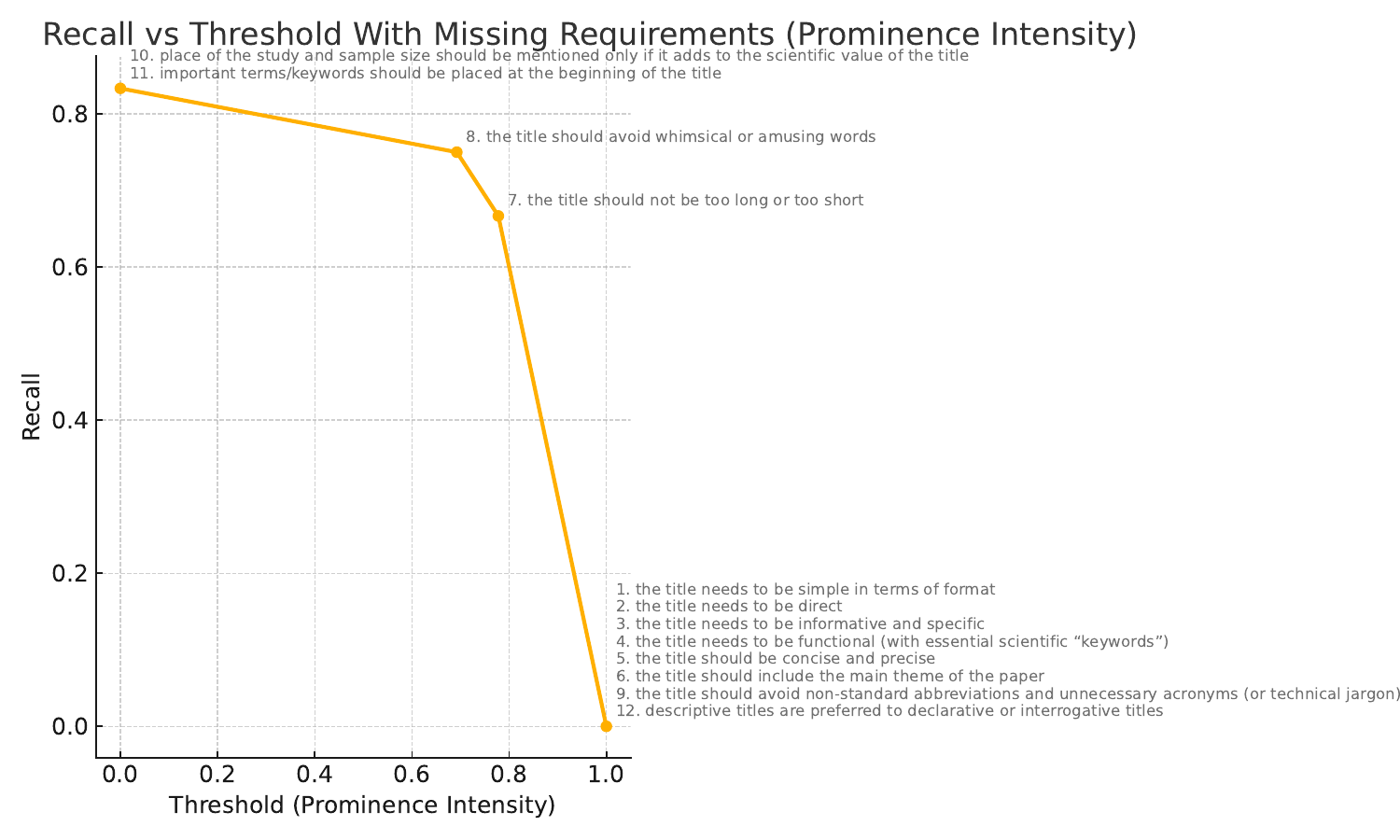}
    \caption{Dropped criterion as Prominence threshold increases.}
    \label{fig:subfig3}
\end{figure*}

The majority of features are retained even at a high threshold of 0.692, suggesting that our methodology is robust in covering essential features. Analysis of the exclusion pattern demonstrates a general relationship between feature importance and retention, although some show divergence from their importance categorization:

\begin{enumerate}
    \item \textbf{Early Exclusions (<0.6)}: None.
    \item \textbf{Mid-Level Exclusions (<0.8)}: Style-related, moderate-importance features including \textit{7. Not too long or short} and \textit{8. Avoid whimsical words} and  are excluded at moderate thresholds.
    \item \textbf{Late (No) Exclusions}: Features of higher importance related to core message delivery and readability remain intact until the highest thresholds, underscoring their fundamental nature. Notable exceptions are \textit{9. Avoid non-standard abbreviations} and \textit{12. Descriptive type}, which are retained despite their mid to lower importance.
\end{enumerate}

\subsection{Frequency Threshold Analysis}

We observe recall changes across varying `Frequency` thresholds.
\begin{table}[h!]
    \centering
    
    \resizebox{\columnwidth}{!}{
    \begin{tabular}{r|r|r}
        \Xhline{0.4mm}
        \textbf{Recall} & \textbf{Threshold (Frequency)} & \textbf{Data Size} \\
         \hline \hline
        0.833 & 1 & 284 \\
        0.750 & 791 & 44 \\
        0.667 & 969 & 32 \\
        0.583 & 1,233 & 26 \\
        0.500 & 1,294 & 20 \\
        0.417 & 1,296 & 19 \\
        0.333 & 1,326 & 13 \\
        0.250 & 1,333 & 11 \\
        0.167 & 1,344 & 4 \\
        0.083 & 1,345 & 1 \\
        \Xhline{0.4mm}
        
    \end{tabular}}
    \caption{Recall and Data Size Across Frequency Thresholds}
\end{table}

Recall remains stable until a threshold of 791. 
The relationship between theoretical importance and retention patterns again reveals some unexpected deviations: a high-importance feature drops out early, while other features of similar importance persist until high thresholds. Additionally, while some supplementary features are not recalled at all, the recalled one (\textit{12. Descriptive type}) shows remarkably strong retention, suggesting that practical title construction may prioritize certain features differently from theoretical guidelines. 
\begin{figure*}[t]
  \centering
    \includegraphics[width=\textwidth]{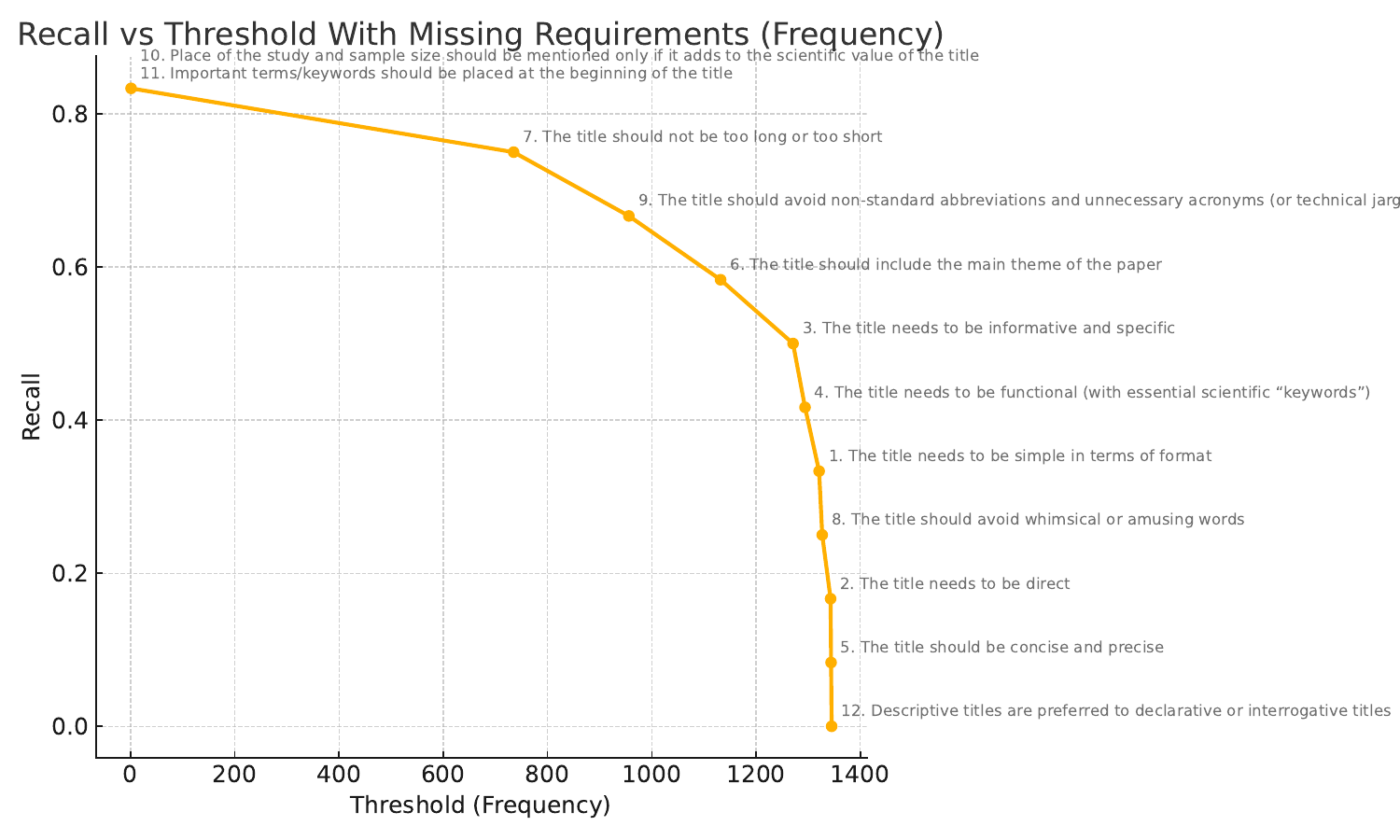}
    \caption{Dropped criterion as Frequency threshold increases}

  \label{title_threshold}
\end{figure*}

\begin{enumerate}
    \item \textbf{Early Exclusions (<500)}: None.
    \item \textbf{Mid-Level Exclusions (<1000)}: Features ranging from moderate (\textit{9. Avoid non-standard abbreviations}) to critical importance (\textit{6. Include the main theme}) are excluded at this stage.
    \item \textbf{Late Exclusions}: Structural features like \textit{1. Simple format} (high importance), \textit{5. Concise and precise} (high importance) persist until the highest thresholds, aligning with their theoretical importance. Interestingly, \textit{12. Descriptive type} (low importance) shows the highest retention despite its low theoretical importance.
\end{enumerate}


\section{DSAI-Generated Top/Bottom Prominence Features of Expert-Driven Annotation Dataset }
\label{appendix:Expert_generated_features}
As discussed in Section \ref{recall}, the features generated using the slogan and title datasets demonstrate high recall values when compared to expert-defined requirements. This holds true even when applying a high threshold for prominence, indicating that the generated features effectively capture the key characteristics outlined by experts.
However, as highlighted in Section \ref{precision}, while features with high prominence exhibit strong discriminative power and reliability, those with lower prominence tend to have relatively lower reliability. 

To provide further insights, we present the top and bottom 20 features for each dataset along with their prominence scores and frequencies. The results for the slogan dataset are detailed in Table \ref{tab:slogan_top_bottom_requirements}, and those for the title dataset are outlined in Table \ref{tab:title_top_bottom_requirements}.

\section{DSAI-Generated Top/Bottom Prominence Features of Industry Dataset}
\label{appendix:Industry_generated_features}

\subsection{Prominence Distribution}
\begin{figure}[t]
\includegraphics[width=\columnwidth]{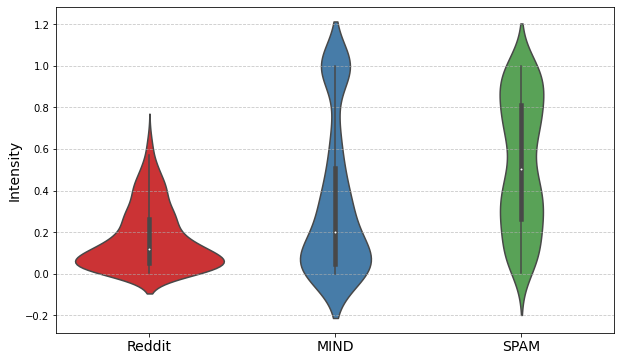}
  \caption{Distribution of each dataset based on prominence scores.}
  \label{fig:industy_dist}
\end{figure}

The following analysis evaluates the prominence score distribution of criteria across three datasets: Reddit, MIND, and SPAM. These scores indicate how reliably each feature identifies the latent characteristics within the datasets.

\paragraph{MIND} The MIND dataset shows a highly concentrated prominence distribution, with most scores reaching 1.0. Few features fall below 0.85, reflecting strong alignment between the generated features and the structured nature of the dataset. These consistently high scores validate the methodology's effectiveness in extracting meaningful patterns from structured data with CTR rates. 

\paragraph{Spam} The Spam detection dataset exhibits a broader prominence distribution than MIND. Still, a large portion are above 0.9, indicating that the methodology is generally effective for SPAM. However, a small subset scoring below 0.5 indicates features that may be less suitable for classification purposes, suggesting the need for careful filtering of specific features for robust classification.

\paragraph{Reddit} The Reddit dataset displays a broad and uneven distribution of prominence scores, peaking at 0.67 with a gradual decline toward lower scores. A significant number of features cluster below 0.4, with many falling below 0.1. This distribution reflects the unstructured and highly diverse nature of Reddit content, making consistent pattern identification challenging. The broad range of topics and content variability suggests that individual features may be highly specific to certain data subsets while falling to generalize across others.

\subsection{General vs Specific Features}

\begin{table*}[t]
\centering
\resizebox{0.7\textwidth}{!}{
\begin{tabular}{l|l|l|r}
\Xhline{0.4mm}
\textbf{Perspective}       & \textbf{Label}                   & \textbf{Type}                    & \textbf{Frequency} \\ 
\Xhline{0.4mm}
\multicolumn{4}{l}{\textbf{General Features}} \\ \Xhline{0.4mm}
reasoning           & Lack of Logical Reasoning                        &    NEGATIVE      & 1,038                \\
fictional\_reference       & Absent                           & NEGATIVE                          & 1,033                \\
gender\_roles              & no\_reference                   & POSITIVE                          & 1,014                \\
archaic\_reference         & no\_reference                   & NEGATIVE                          & 1,004                \\
euphemistic\_language      & euphemism\_absent               & NEGATIVE                          & 986                 \\
historical\_reflection     & Absent                           & NEGATIVE                          & 966                 \\
partisan\_tone             & absent                           & POSITIVE                          & 964                 \\
legal\_context             & Absence                          & POSITIVE                          & 962                 \\
objective\_evaluation      & Subjective                       & NEGATIVE                          & 950                 \\
validity                   & subjective                      & NEGATIVE                          & 926                 \\
scientific\_reference      & No Scientific Reference          & NEGATIVE                          & 918                 \\
profanity                  & Profanity Absent                 & POSITIVE                          & 910                 \\ 
\Xhline{0.4mm}
\multicolumn{4}{l}{\textbf{Specific Features}} \\\Xhline{0.4mm}
specific\_undertone          &      Sarcastic  &             NEGATIVE            & 116                 \\
theme\_recognition         & Justice and Morality            & NEGATIVE                          & 116                 \\
cultural\_sensitivity      & high\_sensitivity               & POSITIVE                          & 114                 \\
efficacy                   & Ineffectiveness                 & POSITIVE                          & 114                 \\
comparison                 & Indirect Comparison             & NEGATIVE                          & 114                 \\
pragmatic                  & Criticism                       & NEGATIVE                          & 110                 \\
audience\_engagement       & Indirect Engagement             & POSITIVE                          & 110                 \\
irony                      & Intensity\_Irony                & NEGATIVE                          & 110                 \\
conflict                   & Presence                        & POSITIVE                          & 104                 \\
demographic\_target        & Youth                           & NEGATIVE                          & 102                 \\
emotionally\_charged       & Emotionally Intense             & NEGATIVE                          & 102                 \\
length                     & two\_digit\_high                & POSITIVE                          & 102                 \\ 
\Xhline{0.4mm}
\end{tabular}}
\caption{Features and their corresponding labels, types, and frequency.}
\label{tab:general_vs_specific}
\end{table*}

DSAI-generated features capture both general characteristics shared across the dataset and highly specific features. As shown in Table \ref{tab:general_vs_specific}, some features represent broad, overarching traits that are present across a significant portion of the dataset, with high frequencies (900+ instances).
These include common patterns such as "lack of logical reasoning" or "absence of historical reflection," which are applicable across diverse contexts. 

In contrast, more specific features capture nuanced and detailed attributes that apply to smaller data subsets, appearing in fewer samples (\~100 instances), such as "sarcastic undertone" or "emotionally intense". These characteristics necessitate fine-grained, data-driven analysis and are typically challenging for LLMs to identify due to their limited presence in pre-training data and the source dataset.
This capability to extract such specific patterns beyond general trends demonstrate DSAI's unique strength over direct feature extraction from LLMs.

The top and bottom 20 prominence fearures for MIND, SPAM and Reddit dataset are provided in Table \ref{tab:mind_top_bottom_requirements}, \ref{tab:spam_top_bottom_requirements}, \ref{tab:reddit_top_bottom_requirements}, respectively.

\begin{table*}[t]
\centering
\resizebox{\textwidth}{!}{
\begin{tabular}{l|r|r}
\Xhline{1.5\arrayrulewidth}
\textbf{Requirement} & \textbf{Prominence} & \textbf{Frequency} \\
\Xhline{1.5\arrayrulewidth}
\multicolumn{3}{c}{\textbf{Top Requirements}} \\
\Xhline{1.5\arrayrulewidth}
The advertising tone should convey a focus on quality, using optimistic and aspirational language. & 0.8571 & 14 \\
The sentence should convey an optimistic advertising tone that encourages engagement. & 0.8333 & 108 \\
The sentence should incorporate cultural references that align with consumptive themes. & 0.8333 & 12 \\
The sentence should employ indirect methods to engage the audience effectively. & 0.7857 & 28 \\
Ensure the sentence contains a component that emotionally appeals to the reader. & 0.7831 & 83 \\
The sentence should effectively incorporate figurative language to enhance meaning. & 0.7556 & 90 \\
The sentence should convey a high level of prominence using strong and emotive language. & 0.7215 & 79 \\
Ensure the sentence conveys its message concisely, avoiding unnecessary words or lengthy expressions. & 0.7143 & 14 \\
The sentence should clearly present a promise of quality, assuring trust and excellence. & 0.6406 & 217 \\
The sentence should employ direct engagement techniques to capture the audience's attention. & 0.6250 & 128 \\
The sentence should avoid losing its identity by being overly generic or broad in purpose. & 0.6190 & 21 \\
Ensure grammatical accuracy throughout the sentence. & 0.6154 & 26 \\
Avoid using the ampersand (\&) in formal writing or titles. & 0.5882 & 34 \\
The sentence should clearly define and communicate a distinct value proposition. & 0.5870 & 184 \\
Ensure the sentence effectively conveys its intended message. & 0.5789 & 19 \\
The sentence should convey a strong and clear emotional tone. & 0.5709 & 275 \\
The sentence should provoke thought and engage the audience in a meaningful way. & 0.5556 & 18 \\
The sentence should avoid presenting the main idea in a way that misaligns with the intended topic or domain. & 0.5556 & 18 \\
The sentence should use persuasive language to effectively encourage action or belief. & 0.5528 & 407 \\
Ensure the sentence includes the mention of a relevant company or organization. & 0.5493 & 213 \\
\Xhline{1.5\arrayrulewidth}
\multicolumn{3}{c}{\textbf{Bottom Requirements}} \\
\Xhline{1.5\arrayrulewidth}
The sentence should include a clear and compelling call to action. & 0.0294 & 1,088 \\
The sentence should fully and effectively communicate its intended message. & 0.0267 & 1,159 \\
The sentence should avoid merely providing information without an intended action or emotion. & 0.0250 & 1,122 \\
Ensure the sentence includes references to cultural significance. & 0.0248 & 1,009 \\
Avoid using imperative or overly complex grammatical structures in titles. & 0.0244 & 41 \\
Ensure the use of inclusive language in the sentence. & 0.0225 & 1,109 \\
Avoid sentences that lack necessary cultural references. & 0.0224 & 1,115 \\
The sentence should not be overly specific, limiting its relevance to a narrow audience. & 0.0205 & 537 \\
The sentence should maintain a general level of specificity to appeal to a wide audience. & 0.0166 & 661 \\
Ensure that sentences maintain grammatical correctness. & 0.0162 & 1,171 \\
The sentence should focus on delivering specific and detailed information. & 0.0148 & 741 \\
Avoid using general language and focus on providing specific details in the sentence. & 0.0131 & 458 \\
The sentence should be concise, conveying its message briefly and efficiently. & 0.0110 & 1,185 \\
The sentence should maintain clear and direct language. & 0.0095 & 1,157 \\
The sentence should avoid overly focusing on the business aspect at the expense of other dimensions. & 0.0070 & 572 \\
The sentence should explicitly mention the presence of a specific service being offered. & 0.0048 & 1,043 \\
The sentence should effectively convey clear and useful information. & 0.0043 & 1,163 \\
The sentence should maintain simplicity in structure and composition. & 0.0034 & ,192 \\
The sentence should be composed in a single language for consistency. & 0.0033 & 1,196 \\
Avoid emotionally neutral language that fails to connect with readers on an emotional level. & 0.0032 & 315 \\
\Xhline{1.5\arrayrulewidth}
\end{tabular}}
\caption{Top and Bottom Requirements of Slogan Dataset based on Prominence.}
\label{tab:slogan_top_bottom_requirements}
\end{table*}

\begin{table*}[t]
\centering
\resizebox{\textwidth}{!}{
\begin{tabular}{l|r|r}
\Xhline{1.5\arrayrulewidth}
\textbf{Requirement} & \textbf{Prominence} & \textbf{Frequency} \\
\Xhline{1.5\arrayrulewidth}
\multicolumn{3}{c}{\textbf{Top Requirements}} \\
\Xhline{1.5\arrayrulewidth}
Avoid using list structures that compromise grammatical integrity and clarity. & 1.0000 & 11 \\
Avoid vague and nonspecific language in the sentence. & 1.0000 & 11 \\
Avoid the use of pronominal subjects in sentences. & 1.0000 & 12 \\
Ensure the sentence provides a clearer focus on the topic. & 1.0000 & 16 \\
Avoid using an informal narrative style in the sentence. & 1.0000 & 33 \\
Ensure the sentence maintains a strong technical focus without shifting to a non-technical direction. & 1.0000 & 41 \\
Ensure the inclusion of phrasal nouns in the sentence where appropriate. & 0.9333 & 30 \\
The sentence should clearly convey its purpose or intent. & 0.9259 & 27 \\
Ensure the sentence addresses the topic clearly without any ambiguity. & 0.8854 & 192 \\
Ensure the sentence clearly specifies its domain of application. & 0.8750 & 16 \\
The sentence should incorporate technical adjectives to enhance precision and clarity in describing nouns or pronouns. & 0.8333 & 12 \\
The sentence should clearly highlight learning methods as the main topic. & 0.8182 & 11 \\
Avoid using interrogative sentences. & 0.8182 & 33 \\
Ensure complete grammatical structures in the sentence. & 0.8095 & 21 \\
Avoid excessive use of concrete nouns in the sentence. & 0.7857 & 28 \\
Avoid overly complex or frequent use of interrogative structures in the content. & 0.7838 & 37 \\
The sentence should avoid being overly verbose and aim for syntactic compression. & 0.7739 & 115 \\
Ensure sentences are concise and avoid unnecessary length. & 0.7500 & 40 \\
The sentence should clearly use either passive or active voice to avoid ambiguity. & 0.7333 & 30 \\
Avoid focusing on concrete, tangible items in the sentence. & 0.7000 & 40 \\
\Xhline{1.5\arrayrulewidth}
\multicolumn{3}{c}{\textbf{Bottom Requirements}} \\
\Xhline{1.5\arrayrulewidth}
Ensure modal verbs are absent in the sentence if not necessary for conveying meaning. & 0.0840 & 917 \\
Limit the length of main noun phrases to enhance readability. & 0.0833 & 96 \\
The main noun phrase in the sentence should ideally consist of two words. & 0.0833 & 144 \\
Avoid using sentence fragments to ensure complete and meaningful sentences. & 0.0805 & 174 \\
Avoid using definite articles when introducing new or less familiar concepts. & 0.0748 & 147 \\
The sentence should effectively use past tense to communicate events or actions that have already occurred. & 0.0667 & 15 \\
The sentence should embrace high complexity in its structure and vocabulary. & 0.0581 & 172 \\
Avoid sentences with overly simplistic structures or vocabulary. & 0.0575 & 435 \\
Ensure sentences utilize participles effectively to enhance clarity and detail. & 0.0529 & 473 \\
Ensure the sentence includes infinitive verb forms for clarity and action orientation. & 0.0526 & 57 \\
Avoid reliance on non-verbal tense constructs in the sentence. & 0.0500 & 40 \\
The sentence should clearly identify and be applicable to the educational domain. & 0.0476 & 21 \\
The sentence should maintain a neutral tone by balancing between passive and active voice. & 0.0476 & 84 \\
Limit the use of multiple verbs in a single sentence. & 0.0455 & 88 \\
Avoid sentences that inappropriately mix technical terms outside the humanities and social sciences domain. & 0.0400 & 25 \\
Limit the use of multiple adjectives in a sentence to maintain clarity and precision. & 0.0357 & 56 \\
The sentence should clearly articulate its relationship to the field of Health and Medicine. & 0.0323 & 31 \\
Ensure the sentence includes well-chosen adjectives to enhance description and clarity. & 0.0121 & 330 \\
Ensure the presence of a clear tense in the sentence. & 0.0076 & 264 \\
Ensure the presence of finite verbs to convey clear action or timing in the sentence. & 0.0039 & 510 \\
\Xhline{1.5\arrayrulewidth}
\end{tabular}}
\caption{Top and Bottom Requirements of Title Dataset based on Prominence.}
\label{tab:title_top_bottom_requirements}
\end{table*}

\begin{table*}[t]
\centering
\resizebox{\textwidth}{!}{
\begin{tabular}{l|r|r}
\Xhline{1.5\arrayrulewidth}
\textbf{Requirement} & \textbf{Prominence} & \textbf{Frequency} \\
\Xhline{1.5\arrayrulewidth}
\multicolumn{3}{c}{\textbf{Top Requirements}} \\
\Xhline{1.5\arrayrulewidth}
The sentence should effectively convey a sense of distress through emotional language. & 1.0000 & 12 \\
Avoid language that implies assistance or support for actions that should be independent. & 0.8462 & 13 \\
Avoid statements that may have an unintended negative impact on the audience. & 0.7895 & 19 \\
Avoid overloading the sentence with complex or unnecessary details related to technology and cybersecurity. & 0.7241 & 29 \\
The sentence should effectively convey a sense of frustration through emotional language. & 0.7143 & 14 \\
The sentence should aim to surprise or alarm the audience to elicit a strong reaction. & 0.6667 & 24 \\
Avoid using language that only implies future outcomes without clear details. & 0.6667 & 18 \\
Avoid using domain-specific language related to the military and defense. & 0.6364 & 11 \\
The sentence should convey information of lesser importance effectively. & 0.6364 & 11 \\
The sentence should avoid focusing solely on technology and innovation when identifying the audience. & 0.6364 & 11 \\
The sentence should clearly specify the legal or criminal aspects of a particular field. & 0.6000 & 35 \\
Avoid using language that fails to motivate or inspire the audience. & 0.5625 & 32 \\
Ensure the main subject of the sentence pertains to sports. & 0.5556 & 18 \\
The sentence should clearly reference past legal cases using appropriate legal terminology. & 0.5385 & 26 \\
Avoid using quality-based adjectives that may imply subjective judgment or bias. & 0.5000 & 24 \\
The sentence should not overly highlight negative issues that could alarm the audience. & 0.5000 & 12 \\
Avoid referencing military actions, personnel, or events in the content. & 0.4737 & 38 \\
The sentence should avoid focusing solely on social and lifestyle aspects within its domain. & 0.4737 & 19 \\
The sentence should avoid using action verbs that emphasize creation-related actions. & 0.4545 & 11 \\
Avoid using ambiguous modal verbs when expressing plans or intentions. & 0.4444 & 18 \\
\Xhline{1.5\arrayrulewidth}
\multicolumn{3}{c}{\textbf{Bottom Requirements}} \\
\Xhline{1.5\arrayrulewidth}
Ensure the inclusion of specific numerical data or statistics in the sentence. & 0.0078 & 766 \\
Avoid overly clear or explicit language that might be inappropriate or too revealing in certain contexts. & 0.0073 & 961 \\
Ensure the sentence does not lack attention to potential contradictions. & 0.0071 & 989 \\
The sentence should exclude references to social media platforms. & 0.0062 & 966 \\
The sentence should effectively communicate its message without relying on figurative language. & 0.0059 & 678 \\
The sentence should avoid the use of separators such as dashes, colons, or slashes. & 0.0054 & 742 \\
Ensure the inclusion of sensory details in the sentence to enhance vividness. & 0.0051 & 973 \\
The sentence should clearly provide instruction or guidance to the reader. & 0.0051 & 979 \\
The sentence should present concrete, specific ideas and details. & 0.0042 & 946 \\
Ensure that the sentence is clear and needs no further clarification. & 0.0042 & 952 \\
Ensure that scientific theories or principles are present in the sentence. & 0.0041 & 976 \\
The sentence should avoid the use of hashtags. & 0.0040 & 994 \\
The sentence should provide sufficient context to be understood independently. & 0.0035 & 865 \\
Avoid or limit the use of figurative language in the sentence. & 0.0032 & 313 \\
Avoid sentences that lack necessary negations to clarify intended meaning. & 0.0031 & 977 \\
Avoid using ambiguous or unsuitable speech acts in sentence construction. & 0.0030 & 989 \\
The sentence should avoid hypothetical scenarios to maintain factual clarity. & 0.0022 & 898 \\
The sentence should avoid content that lacks relevance or significance, especially for a news context. & 0.0020 & 980 \\
The sentence should clearly identify the language being used, ensuring linguistic features are consistent with the specified language. & 0.0020 & 998 \\
The sentence should be free from unnecessary repetition, ensuring clarity and conciseness. & 0.0010 & 985 \\
\Xhline{1.5\arrayrulewidth}
\end{tabular}}
\caption{Top and Bottom Requirements of MIND dataset based on Prominence.}
\label{tab:mind_top_bottom_requirements}
\end{table*}

\begin{table*}[t]
\centering
\resizebox{\textwidth}{!}{
\begin{tabular}{l|r|r}
\Xhline{1.5\arrayrulewidth}
\textbf{Requirement} & \textbf{Prominence} & \textbf{Frequency} \\
\Xhline{1.5\arrayrulewidth}
\multicolumn{3}{c}{\textbf{Top Requirements}} \\
\Xhline{1.5\arrayrulewidth}
Avoid using list structures that compromise grammatical integrity and clarity. & 1.0000 & 11 \\
Avoid vague and nonspecific language in the sentence. & 1.0000 & 11 \\
Avoid the use of pronominal subjects in sentences. & 1.0000 & 12 \\
Ensure the sentence provides a clearer focus on the topic. & 1.0000 & 16 \\
Avoid using an informal narrative style in the sentence. & 1.0000 & 33 \\
Ensure the sentence maintains a strong technical focus without shifting to a non-technical direction. & 1.0000 & 41 \\
Ensure the inclusion of phrasal nouns in the sentence where appropriate. & 0.9333 & 30 \\
The sentence should clearly convey its purpose or intent. & 0.9259 & 27 \\
Ensure the sentence addresses the topic clearly without any ambiguity. & 0.8854 & 192 \\
Ensure the sentence clearly specifies its domain of application. & 0.8750 & 16 \\
The sentence should incorporate technical adjectives to enhance precision and clarity in describing nouns or pronouns. & 0.8333 & 12 \\
The sentence should clearly highlight learning methods as the main topic. & 0.8182 & 11 \\
Avoid using interrogative sentences. & 0.8182 & 33 \\
Ensure complete grammatical structures in the sentence. & 0.8095 & 21 \\
Avoid excessive use of concrete nouns in the sentence. & 0.7857 & 28 \\
Avoid overly complex or frequent use of interrogative structures in the content. & 0.7838 & 37 \\
The sentence should avoid being overly verbose and aim for syntactic compression. & 0.7739 & 115 \\
Ensure sentences are concise and avoid unnecessary length. & 0.7500 & 40 \\
The sentence should clearly use either passive or active voice to avoid ambiguity. & 0.7333 & 30 \\
Avoid focusing on concrete, tangible items in the sentence. & 0.7000 & 40 \\
\Xhline{1.5\arrayrulewidth}
\multicolumn{3}{c}{\textbf{Bottom Requirements}} \\
\Xhline{1.5\arrayrulewidth}
Ensure modal verbs are absent in the sentence if not necessary for conveying meaning. & 0.0840 & 917 \\
Limit the length of main noun phrases to enhance readability. & 0.0833 & 96 \\
The main noun phrase in the sentence should ideally consist of two words. & 0.0833 & 144 \\
Avoid using sentence fragments to ensure complete and meaningful sentences. & 0.0805 & 174 \\
Avoid using definite articles when introducing new or less familiar concepts. & 0.0748 & 147 \\
The sentence should effectively use past tense to communicate events or actions that have already occurred. & 0.0667 & 15 \\
The sentence should embrace high complexity in its structure and vocabulary. & 0.0581 & 172 \\
Avoid sentences with overly simplistic structures or vocabulary. & 0.0575 & 435 \\
Ensure sentences utilize participles effectively to enhance clarity and detail. & 0.0529 & 473 \\
Ensure the sentence includes infinitive verb forms for clarity and action orientation. & 0.0526 & 57 \\
Avoid reliance on non-verbal tense constructs in the sentence. & 0.0500 & 40 \\
The sentence should clearly identify and be applicable to the educational domain. & 0.0476 & 21 \\
The sentence should maintain a neutral tone by balancing between passive and active voice. & 0.0476 & 84 \\
Limit the use of multiple verbs in a single sentence. & 0.0455 & 88 \\
Avoid sentences that inappropriately mix technical terms outside the humanities and social sciences domain. & 0.0400 & 25 \\
Limit the use of multiple adjectives in a sentence to maintain clarity and precision. & 0.0357 & 56 \\
The sentence should clearly articulate its relationship to the field of Health and Medicine. & 0.0323 & 31 \\
Ensure the sentence includes well-chosen adjectives to enhance description and clarity. & 0.0121 & 330 \\
Ensure the presence of a clear tense in the sentence. & 0.0076 & 264 \\
Ensure the presence of finite verbs to convey clear action or timing in the sentence. & 0.0039 & 510 \\
\Xhline{1.5\arrayrulewidth}
\end{tabular}}
\caption{Top and Bottom Requirements of Spam Detection Dataset based on Prominence.}
\label{tab:spam_top_bottom_requirements}
\end{table*}

\begin{table*}[t]
\centering
\resizebox{\textwidth}{!}{
\begin{tabular}{l|r|r}
\Xhline{1.5\arrayrulewidth}
\textbf{Requirement} & \textbf{Prominence} & \textbf{Frequency} \\
\Xhline{1.5\arrayrulewidth}
\multicolumn{3}{c}{\textbf{Top Requirements}} \\
\Xhline{1.5\arrayrulewidth}
The sentence should avoid unnecessary criticism and focus on constructive feedback. & 0.6727 & 110 \\
The sentence should avoid using strong partisan language or political bias. & 0.6129 & 124 \\
The sentence should include elements that evoke a surreal or dream-like quality. & 0.6104 & 154 \\
The sentence should avoid being perceived solely as a review. & 0.6082 & 194 \\
The sentence should effectively convey a positive emotion. & 0.5699 & 186 \\
The sentence should effectively convey information and facilitate informative sharing with the audience. & 0.5333 & 120 \\
Avoid using a critical tone in the sentence. & 0.5313 & 128 \\
The sentence should convey a positive sentiment. & 0.5294 & 136 \\
Avoid language that inaccurately places the sentence within the politics and media domain. & 0.5238 & 126 \\
Ensure the sentence does not include language indicating complaints or grievances. & 0.5130 & 230 \\
The sentence should align with public opinion and mainstream consensus. & 0.5062 & 162 \\
The sentence should address topics in a non-sensitive manner, avoiding contentious or inflammatory language. & 0.5044 & 226 \\
Avoid displaying a dominant interpersonal stance in relational dynamics. & 0.4907 & 161 \\
The sentence should convey a positive emotional tone. & 0.4815 & 108 \\
The sentence should convey information with high intensity and detail to maximize audience knowledge gain. & 0.4737 & 190 \\
The content should maintain an objective, unbiased perspective throughout. & 0.4729 & 129 \\
The sentence should use words with optimistic or positive connotations. & 0.4516 & 124 \\
The language used should maintain a neutral tone, avoiding extreme politeness or rudeness. & 0.4476 & 210 \\
Avoid using circular or flawed reasoning in arguments. & 0.4468 & 188 \\
The sentence should avoid rhetorical devices and maintain a straightforward approach. & 0.4430 & 237 \\
\Xhline{1.5\arrayrulewidth}
\multicolumn{3}{c}{\textbf{Bottom Requirements}} \\
\Xhline{1.5\arrayrulewidth}
The content should maintain a moderate level of complexity. & 0.0116 & 518 \\
Avoid non-impressionistic language and incorporate more vivid or subjective descriptions. & 0.0110 & 182 \\
The sentence should avoid legal language or references. & 0.0104 & 962 \\
The sentence should avoid suggesting a lack of consensus or collective agreement. & 0.0099 & 202 \\
The sentence should appropriately reference hierarchical authority to establish credibility or significance. & 0.0093 & 214 \\
Avoid using American regional dialects or cultural references in the sentence. & 0.0090 & 222 \\
The sentence should avoid depicting any specific gender roles or influences. & 0.0081 & 867 \\
Ensure the sentence includes relevant jargon where necessary to convey expertise and precision. & 0.0080 & 754 \\
The sentence should avoid overly intense expressions of real-world relevance. & 0.0062 & 1,129 \\
Ensure the sentence length is concise, ideally with a word count in the single digits. & 0.0061 & 330 \\
The sentence should avoid making explicit factual assertions without sufficient evidence or context. & 0.0060 & 668 \\
The sentence should be inclusive and cater to a broad demographic audience. & 0.0056 & 354 \\
The sentence should clearly demonstrate an understanding of various sentence types and structures. & 0.0053 & 1,126 \\
The sentence should not restrict or misrepresent opinion sharing. & 0.0047 & 426 \\
The sentence should be intellectually demanding but still accessible at a medium level of complexity. & 0.0045 & 446 \\
Avoid using exclusive language or sentiments in the sentence. & 0.0043 & 235 \\
Minimize the use of overly dramatic or theatrical language in the sentence. & 0.0027 & 375 \\
The sentence should maintain a concise length, ideally within a low two-digit word count. & 0.0026 & 389 \\
The sentence should intentionally exclude hashtags. & 0.0025 & 1,191 \\
Ensure language identification and classification are accurate within the sentence. & 0.0017 & 1,194 \\
\Xhline{1.5\arrayrulewidth}
\end{tabular}}
\caption{Top and Bottom Requirements of Reddit dataset based on Prominence.}
\label{tab:reddit_top_bottom_requirements}
\end{table*}

\clearpage

\begin{figure*}[h]
\includegraphics[width=\textwidth]{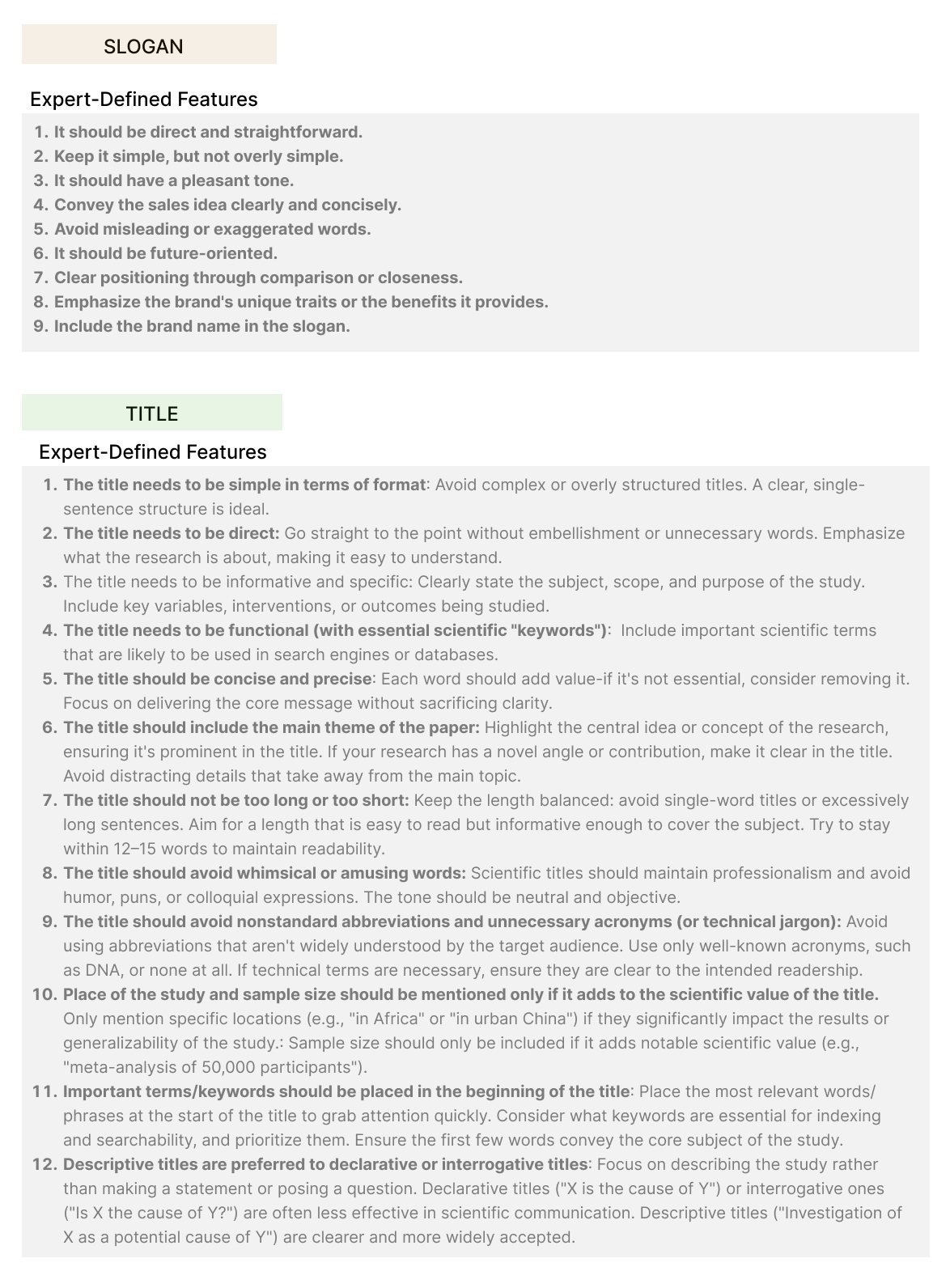}
  \caption{Expert-Defined Criteria}
  \label{expert_defined}
\end{figure*}



\begin{figure*}[t]
\includegraphics[width=\textwidth]{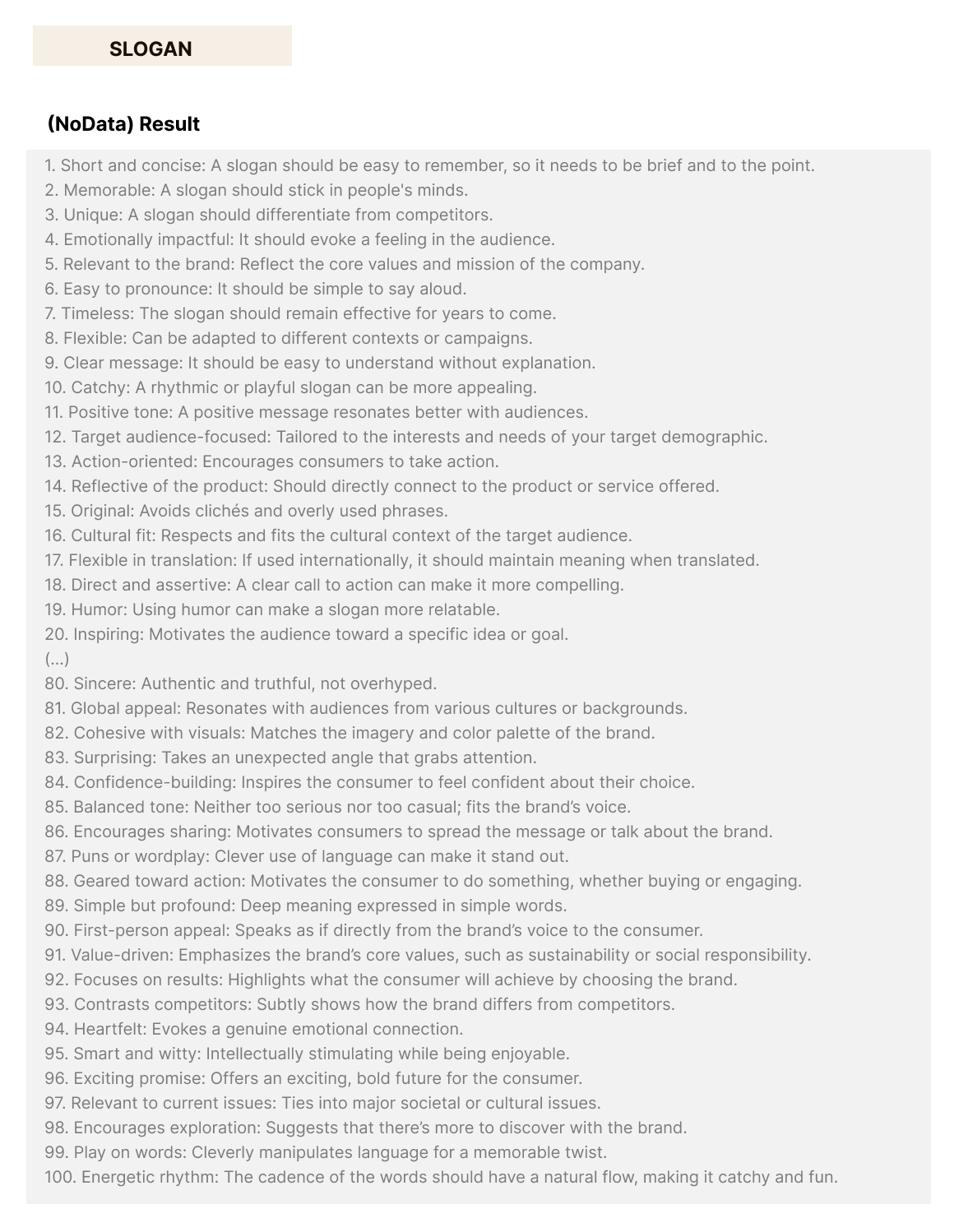}
  \caption{NoData Result for slogan}
  \label{fig:baseline_results}
\end{figure*}

\clearpage
\begin{figure*}[t]
\includegraphics[width=\textwidth]{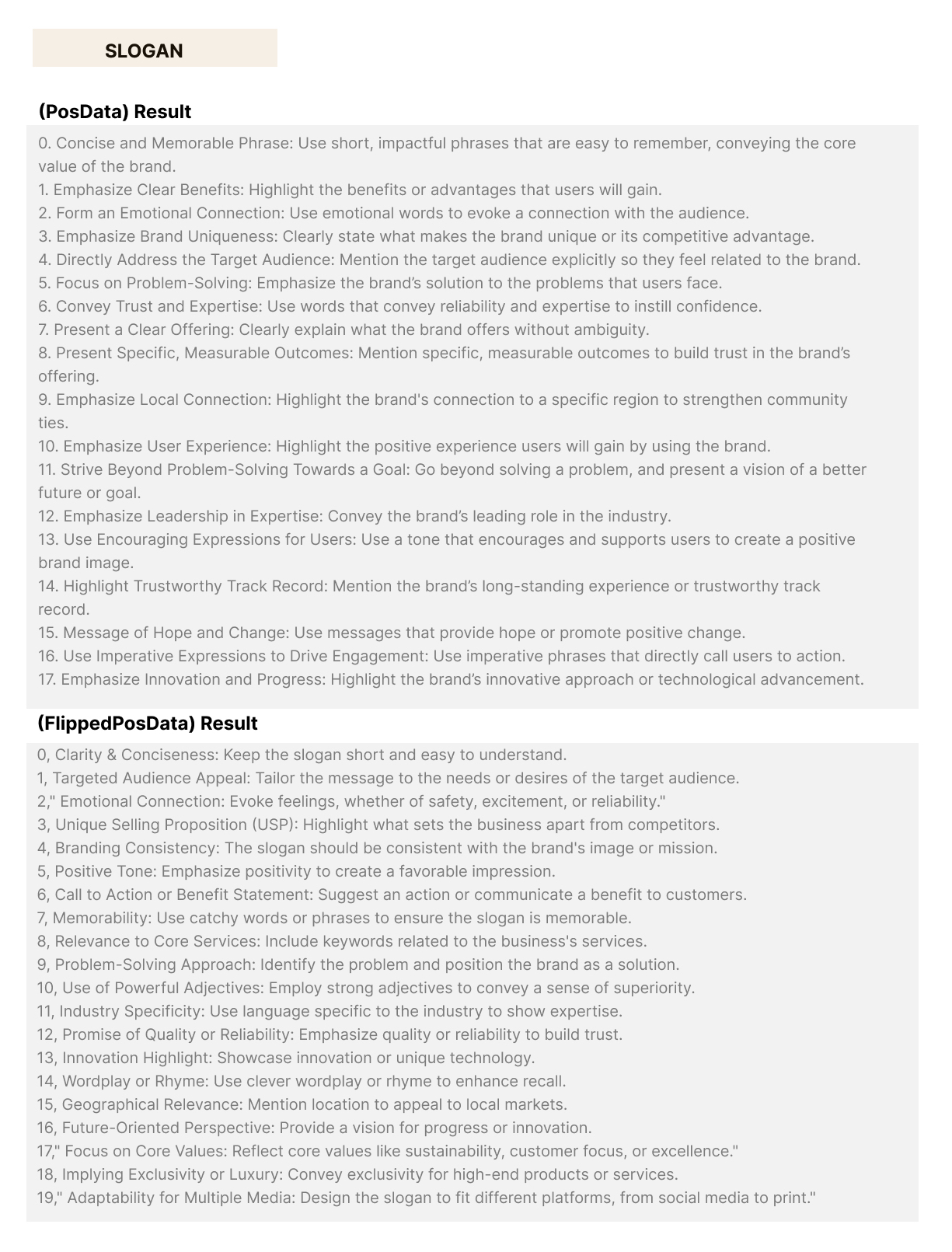}
  \caption{PosData and FlippedPosData Results for slogan}
  \label{fig:baseine2,3_slogan}
\end{figure*}

\clearpage
\begin{figure*}[t]
\includegraphics[width=\textwidth]{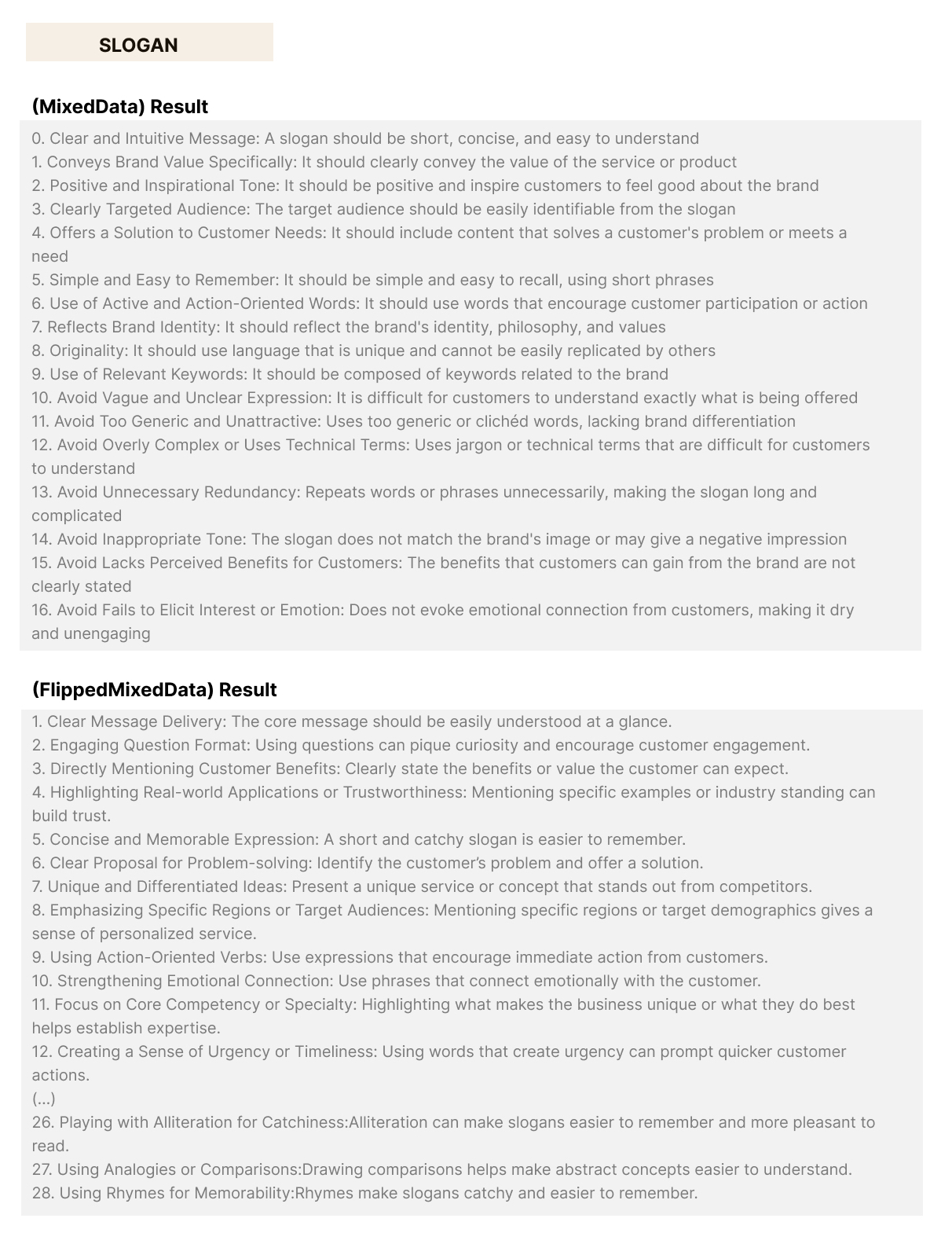}
  \caption{MixedData and FlippedMixedData Results for slogan}
  \label{fig:baseine4,5_slogan}
\end{figure*}

\clearpage
\begin{figure*}[t]
\includegraphics[width=\textwidth]{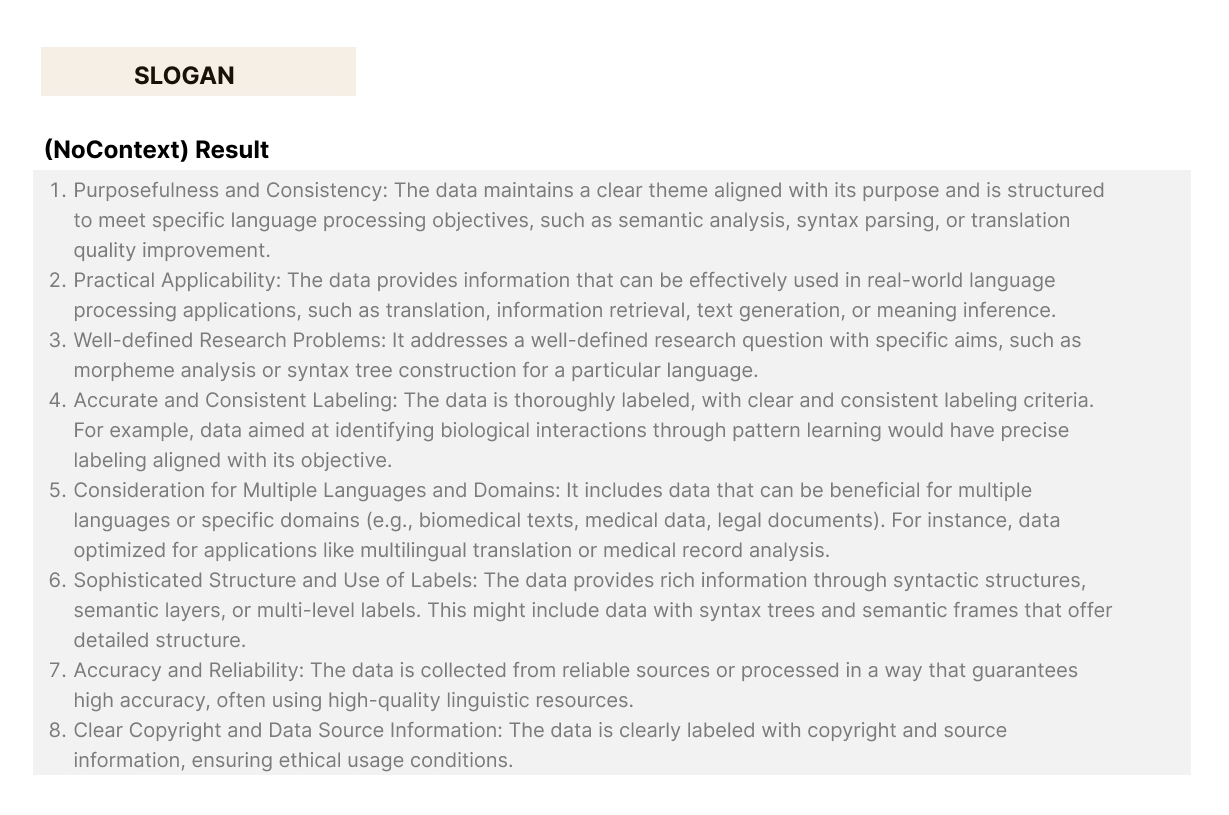}
  \caption{NoContext Result for slogan}
  \label{fig:baseine6_slogan}
\end{figure*}

\clearpage
\begin{figure*}[t]
\includegraphics[width=\textwidth]{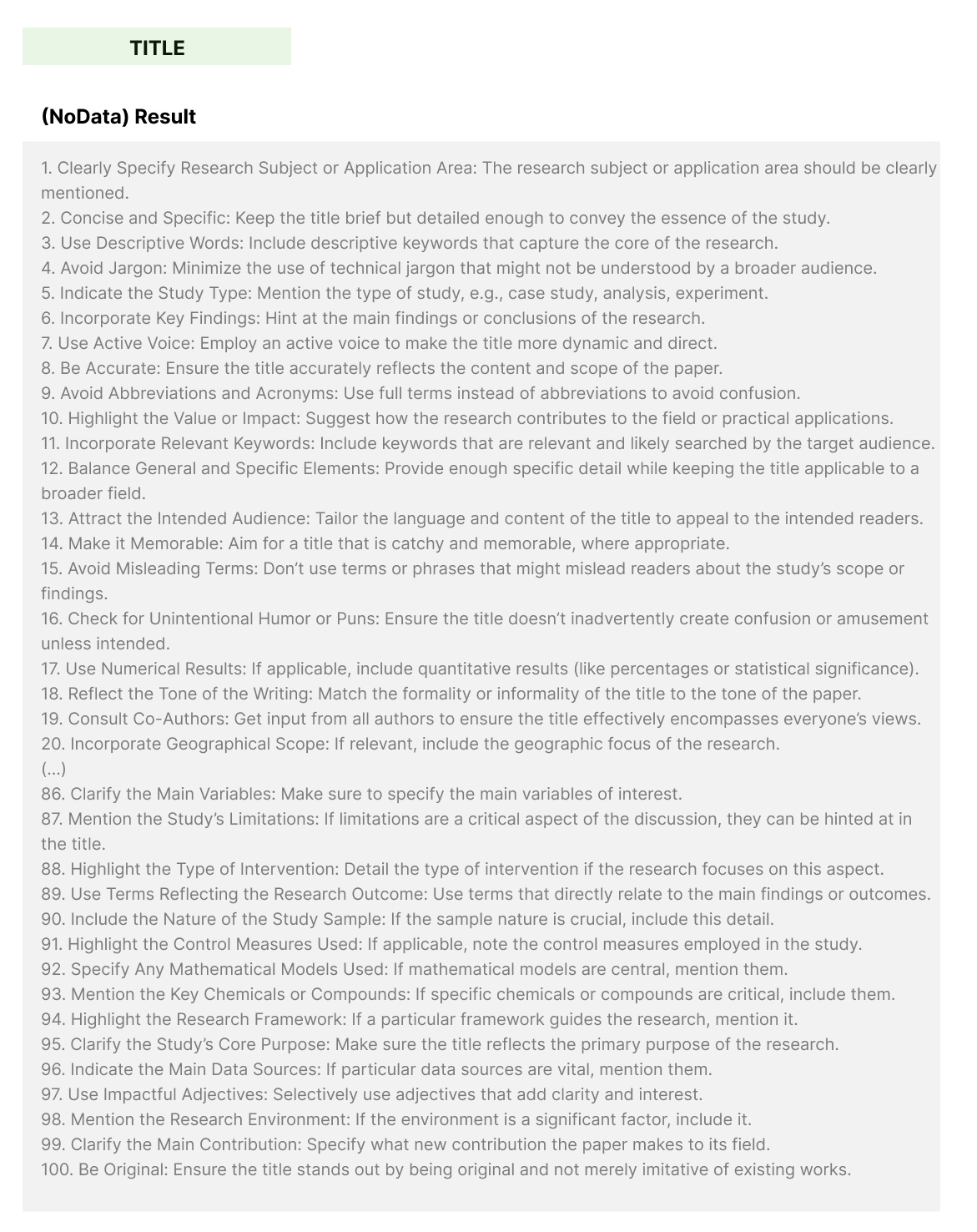}
  \caption{NoData Result for title}
  \label{fig:baseine1_title}
\end{figure*}

\clearpage
\begin{figure*}[t]
\includegraphics[width=\textwidth]{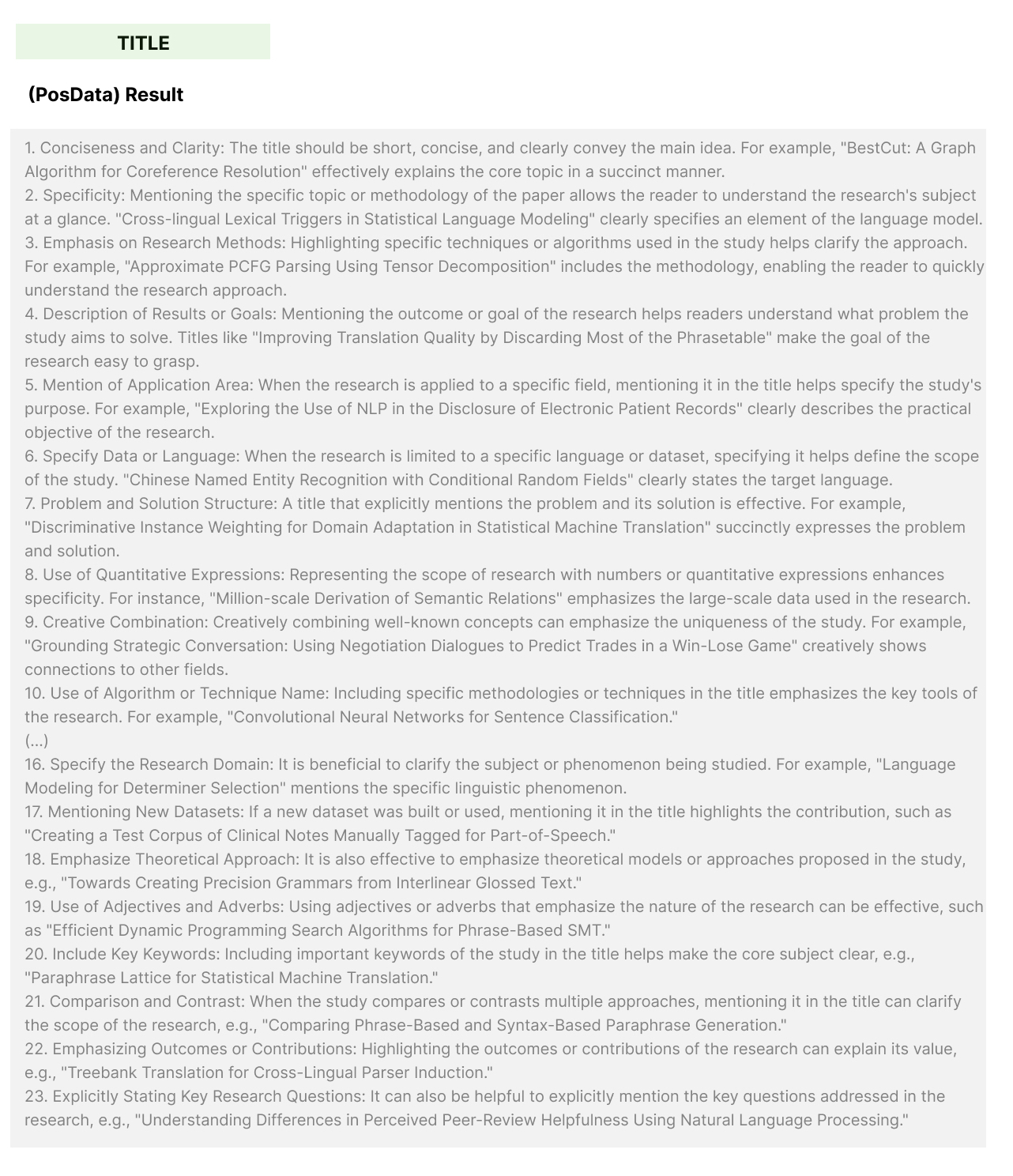}
  \caption{PosData Result for title}
  \label{fig:baseine2_title}
\end{figure*}

\clearpage
\begin{figure*}[t]
\includegraphics[width=\textwidth]{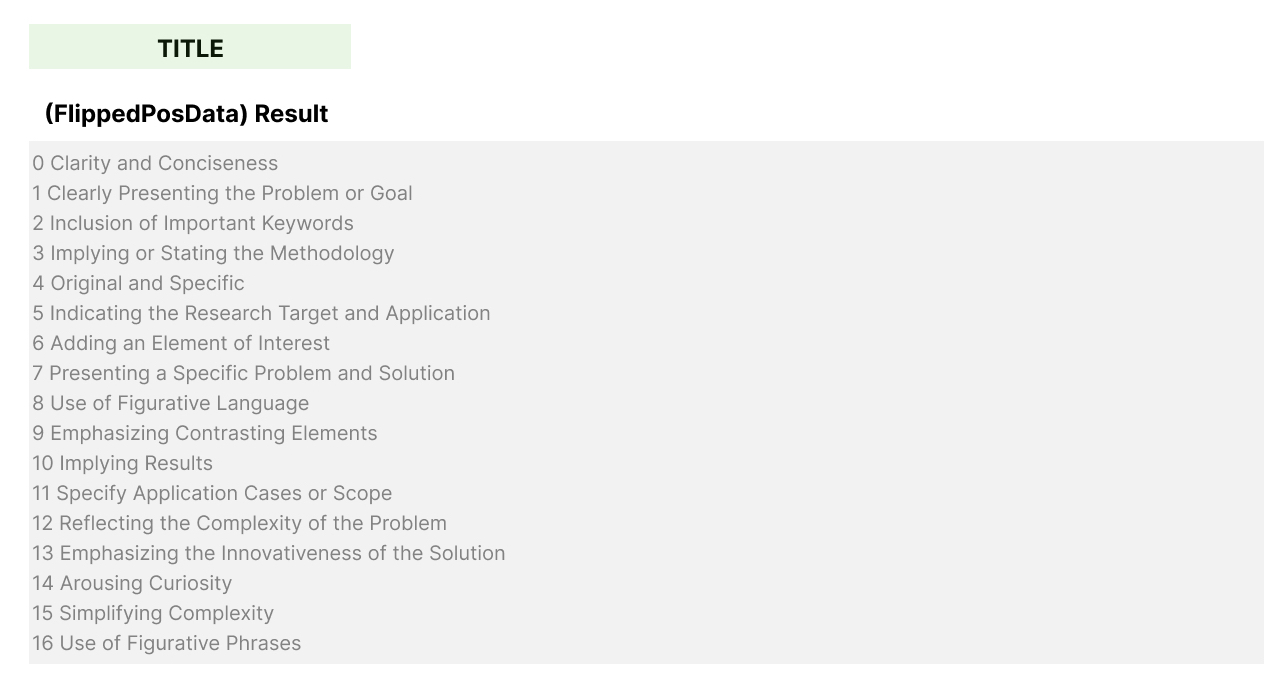}
  \caption{FlippedPosData Result for title}
  \label{fig:baseine3_title}
\end{figure*}

\clearpage
\begin{figure*}[t]
\includegraphics[width=\textwidth]{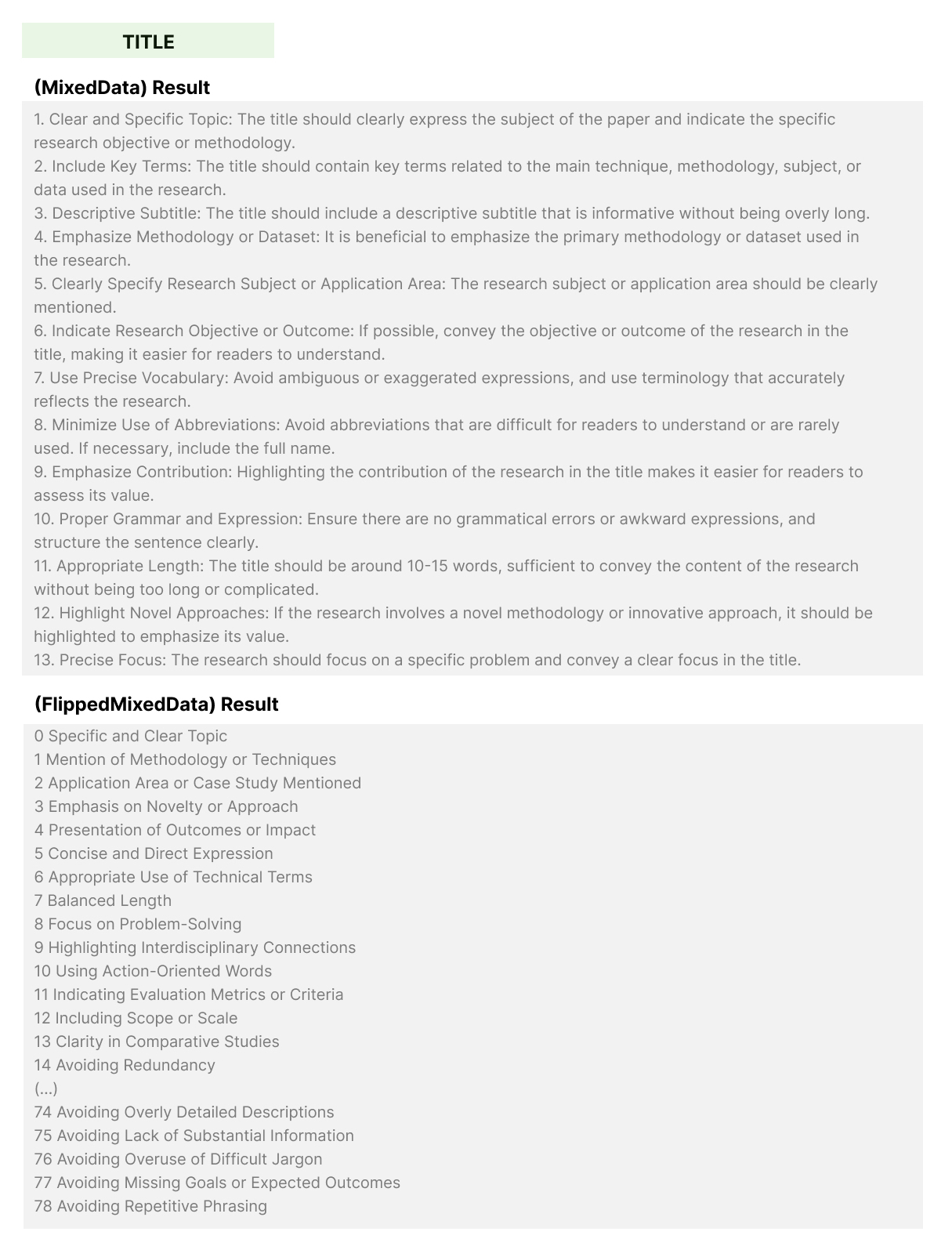}
  \caption{MixedData and FlippedMixedData Results for title}
  \label{fig:baseine4,5_title}
\end{figure*}

\clearpage
\begin{figure*}[t]
\includegraphics[width=\textwidth]{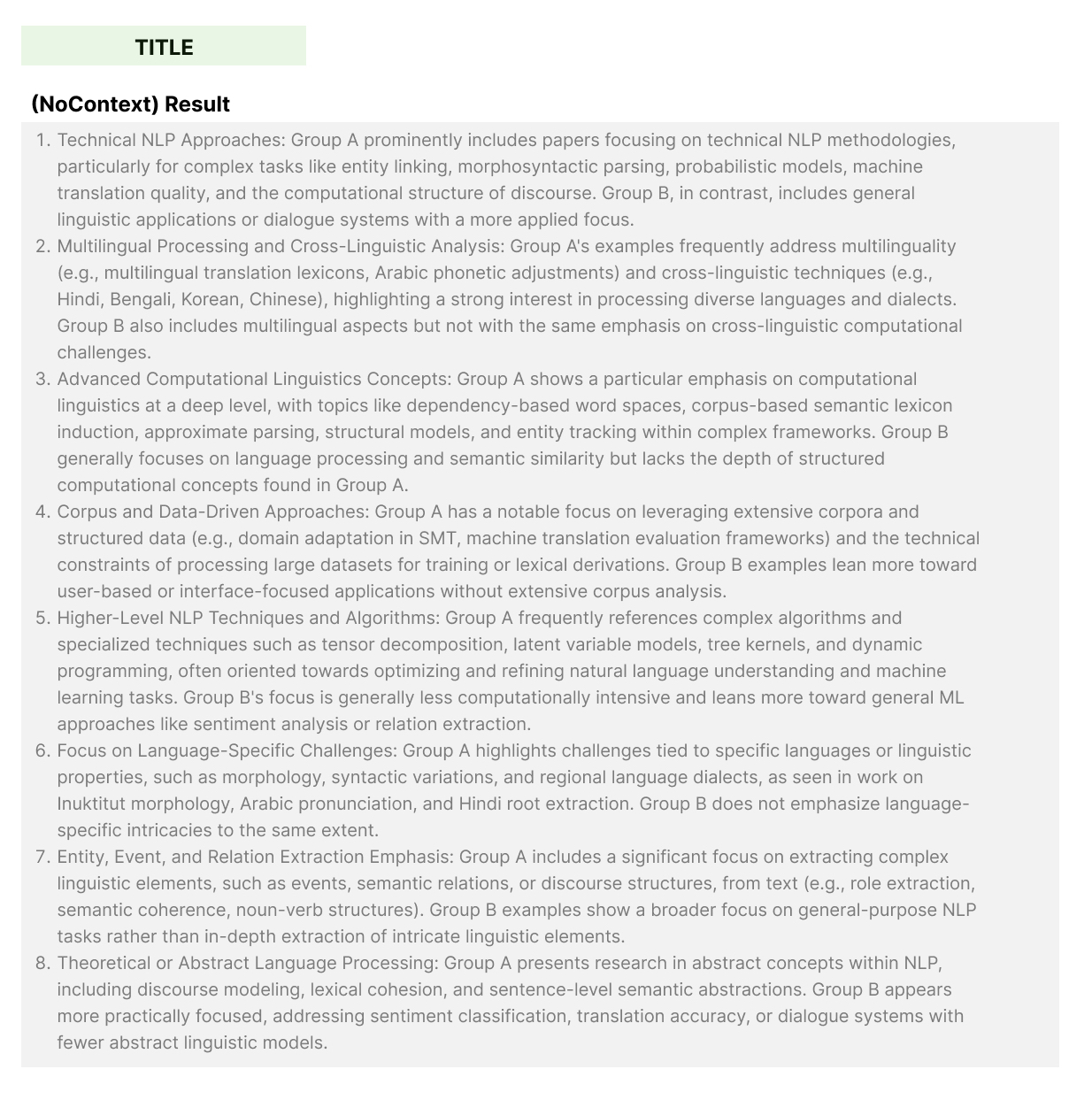}
  \caption{NoContext Result for title}
  \label{fig:baseine6_title}
\end{figure*}
\clearpage

\end{document}